\documentclass[journal]{IEEEtran}
\IEEEoverridecommandlockouts

\usepackage{amsmath, amssymb, amsfonts, mathtools, cases}  
\usepackage{threeparttable, multirow, booktabs}   
\usepackage{enumitem}
\usepackage{graphicx}
\usepackage{bbm, verbatim}
\usepackage[sort]{cite}

\newcommand*{\sysname}{LoRaCompass}
\newcommand{\FigureWord}[0] {Figure}
\newcommand{\FigureWords}[0] {Figures}
\newcommand{\TableWord}[0] {Table}
\newcommand{\TableWords}[0] {Tables}
\newcommand{\EquationWord}[0] {\unskip}
\newcommand{\EquationWords}[0] {\unskip}
\newcommand{\SectionWord}[0] {Section}
\newcommand{\SectionWords}[0] {Sections}
\newcommand{\prob}[0] {P}

\begin{document}

\title{
  \sysname{}: Robust Reinforcement Learning\\ to Search for a LoRa Tag
}

\author{
      Tianlang~He,
      Zhongming~Lin,
      Tianrui~Jiang,
      S.-H. Gary Chan,~\IEEEmembership{Senior Member,~IEEE}
      \thanks{T. He, Z. Lin, T. Jiang, and S.-H. G. Chan are with 
      the Department of Computer Science and Engineering, 
      The Hong Kong University of Science and Technology, Hong Kong 
      (e-mail: theaf@cse.ust.hk, 
      zmhkust@ust.hk, 
      tjiangag@connect.ust.hk,
      gchan@cse.ust.hk; 
      corresponding author: Tianlang He. )}
}

\maketitle

\begin{abstract}

Long-Range (LoRa) communication, known for its extensive range and low power consumption, 
has increasingly been adopted in IoT tags worn by mentally incapacitated persons (MIPs) and others at risk of going missing.
We study how to leverage reinforcement learning to enable
a mobile sensor to search for a periodically broadcasting LoRa tag in unknown environments, 
guided by the received signal strength indicator (RSSI) on the fly. 
Previous approaches to this problem rely on RSSI temporal features, which are vulnerable to signal fluctuation and domain shifts across environments, 
often leading to localization inaccuracies. 
We propose \sysname{}, a zero-shot generalization (ZSG) reinforcement learning model,  
which uses RSSI spatial features to achieve robust and efficient search for a LoRa tag in unknown environments. 
For exploitation, \sysname{} reconstructs RSSI time series into a signal map, 
based on which it maximizes the probability of the LoRa sensor moving closer to a tag. 
For exploration, it uses a UCB-inspired (upper confidence bound) function to maximize the completeness of the signal map, so as to 
guide the sensor with increasing confidence. 
We have validated \sysname{} in ground-based and drone-assisted scenarios within diverse unseen environments that cover areas of over $80\,\mbox{km}^2$ in total.  
\sysname{} shows a high success rate ($\textgreater90\%$) for locating LoRa tags within $100\,\mbox{m}$ (a $40\%$ improvement over existing methods),  
and satisfactory efficiency, where the search path length grows linearly with the initial distance between sensor and tag. 


\end{abstract}

\section{Introduction}\label{sec:intro}

The Long Range (LoRa) protocol has long been used for Internet of Things (IoT) communication, 
known for its wide coverage (1--5km), low power consumption, and low cost~\cite{aldhaheri2024lora}. 
These traits have led to its adoption in IoT tags worn by individuals at risk of going missing, 
such as mentally incapacitated patients (MIPs—including those with dementia, Down syndrome, autism, or psychological disorders), children, and hikers~\cite{quero2025unmanned, zhang2023rf, tong2024robots, manuel2024lora, hu2021distributed, FLCCMC24}. 
Such a {\em LoRa tag} can broadcast signals (roughly once per minute) over kilometers for months on a single battery. 
The longevity and long range provide sustained safety support for those individuals, making LoRa tags 
superior to other IoT devices such as the power-hungry GPS trackers and short-range Bluetooth tags~\cite{he2022tackling, hany2024airtags, hadwen2017energy, sadeghian2021review}. 

When an individual is reported missing, rescuers localize a LoRa tag (and thus the individual) using either fixed LoRa gateways or a {\em mobile sensor}.   
Recently, the mobile scheme has gained increasing traction, since it operates independently of LoRa infrastructure~\cite{li2021urban,shi2024enable}. 
In the mobile scheme, a {\em mobile agent} (e.g., a caregiver, vehicle, or drone) carries a LoRa sensor and moves sequentially toward a tag, 
guided by the received signal strength indicator (RSSI) at the sensor.\footnote{RSSI is easy to obtain and suitable for non-line-of-sight scenarios~\cite{elsherif2025theoretical}.}  
The agent searches for the location with the maximal RSSI, where the tag is normally located. 

\begin{figure}[tp]
  \centering
  \includegraphics[width=0.95\linewidth]{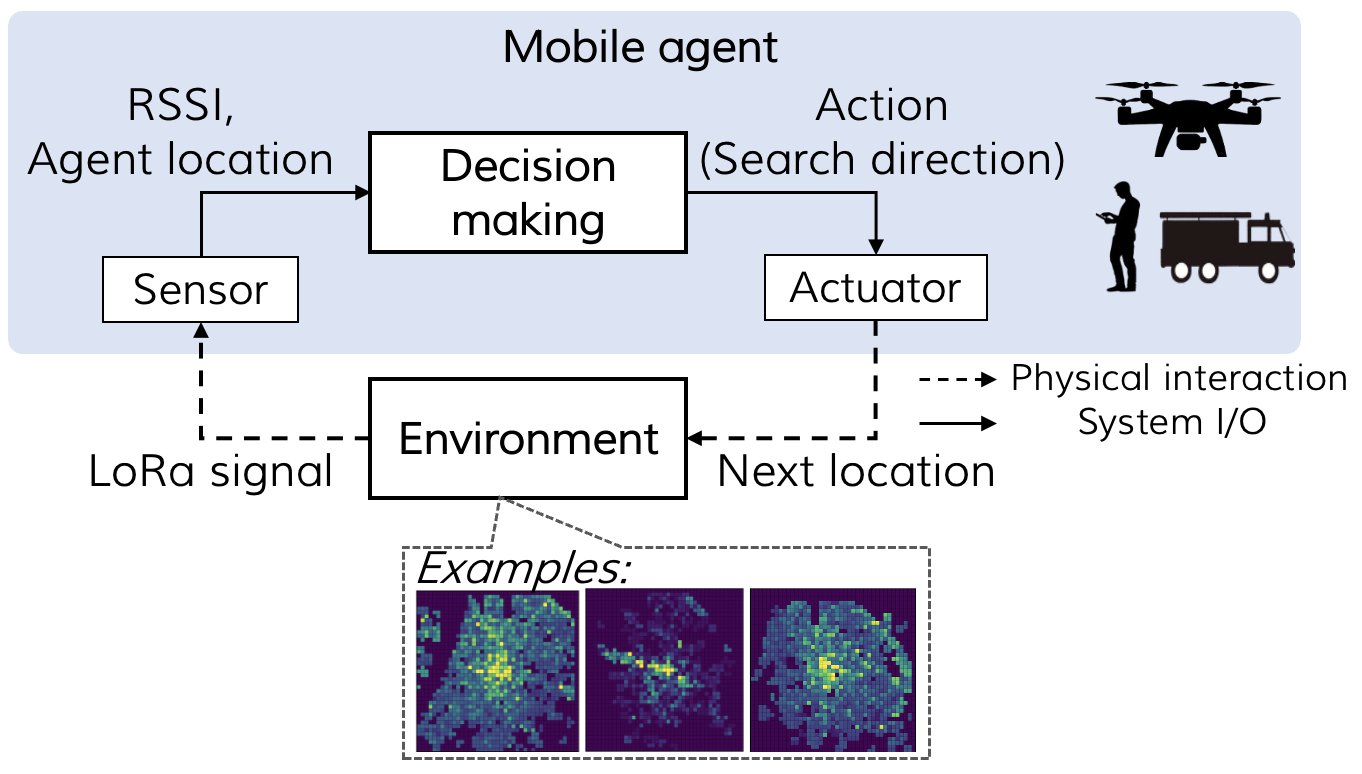}
  \caption{Sequential decision-making process for a mobile agent interacting with an environment
  to localize a LoRa tag. 
    At each step, the agent waits for a LoRa signal and measures the RSSI at its location; then, 
  it decides the action to hop to the next location.  
  This problem is challenging due to the complex real-world environments, 
  each defining a unique mapping from the agent's location to RSSI 
  (the RSSI heatmaps illustrate the different environments, with the LoRa tag placed in the middle of the map). 
  } 
  \label{fig:cycle} 
\end{figure} 

We divide a search area into a set of regular {\em grid points}, 
where the grid size corresponds to the distance an agent can travel during the broadcast interval of the LoRa tag.\footnote{For instance, if an agent moves at a speed of 50 meters per minute, and the tag broadcasts every 2 minutes, the grid size is 100 meters. } 
The agent's movement to a neighboring grid point is defined as a hop, and the direction chosen for a hop is defined as an {\em action}.
\FigureWord~\ref{fig:cycle} illustrates the search process for a mobile agent to localize a LoRa tag.    
At each step, the agent waits for a LoRa signal and measures the RSSI at its current grid point. Then, given the RSSI and grid locations, 
it decides on the action and hops to the next grid point. 
Optimally, the agent always chooses the action that minimize its distance to the tag, namely the {\em tag's direction}. 

We study the sequential action-decision process for a mobile agent to localize a LoRa tag in unknown {\em environments} (urban, rural, forested areas, etc.), 
using only the RSSI measurements at the agent's locations. 
This problem is challenging because the tag direction is difficult to determine when the environment is unknown.  
In unknown environments, various dynamic (e.g., multipath fading, ambient noise) 
and static (e.g., buildings, vegetation, terrain) factors affect LoRa signal propagation, and thus each environment defines a unique mapping from the agent's location to RSSI (shown as signal heatmaps in \FigureWord~\ref{fig:cycle}). 
As a result, agents must infer tag directions under {\em RSSI fluctuation} and {\em domain shift}, 
which challenges the accurate localization of LoRa tags in unknown environments.  

Previous studies have not fully considered the adverse effects of RSSI fluctuation and domain shift on this search problem. 
Due to RSSI fluctuation, an agent at the same location may obtain different RSSI measurements, 
and the same RSSI may be observed at different locations. 
This disturbs previous methods that uses path loss models to determine tag directions, 
as an RSSI increase does not necessarily indicate proximity to the tag~\cite{sorbelli2020measurement, shen2022cloud, toulis2021proximal, gao2012implementing}. 
Furthermore, even without fluctuation, domain shift makes  
the underlying RSSI landscape irregular, unpredictable, and variable across environments. 
Although existing approaches extract the RSSI temporal feature for estimating tag directions via reinforcement learning (RL), 
the feature from one environment often fail to generalize to another, 
leading to cascading decision errors that culminate in localization inaccuracies~\cite{soorki2024catch, zhu2024emergency, gu2020uav, kirk2023survey, ossongo2024multi,moos2022robust, wang2021online}.
Hence, this problem calls for a {\em generalizable} RSSI feature that can robustly represent the tag direction across environments. 

We observe that the RSSI spatial variation within a large area 
provides a robust representation for the tag direction, notwithstanding RSSI fluctuation and domain shift. 
Unlike the local myopic RSSI time series, 
RSSI from a macroscopic perspective (say, over a $20\times 20$ grid) exhibits a clear pattern --- RSSI readings are statistically higher when the grid points are closer to the tag (see examples in Appendix~\ref{sec:appendix_heatmap}). 
Such an {\em RSSI spatial feature} stems from the physical principle of signal attenuation~\cite{mhatre2025signal}, 
which may provide RL with a generalizable inductive bias for the tag direction, robust against environmental factors. 
Therefore, we focus on how to leverage the RSSI spatial feature to enable LoRa tag search across environments. 

\begin{figure}[tp]
  \centering
  \includegraphics[width=\linewidth]{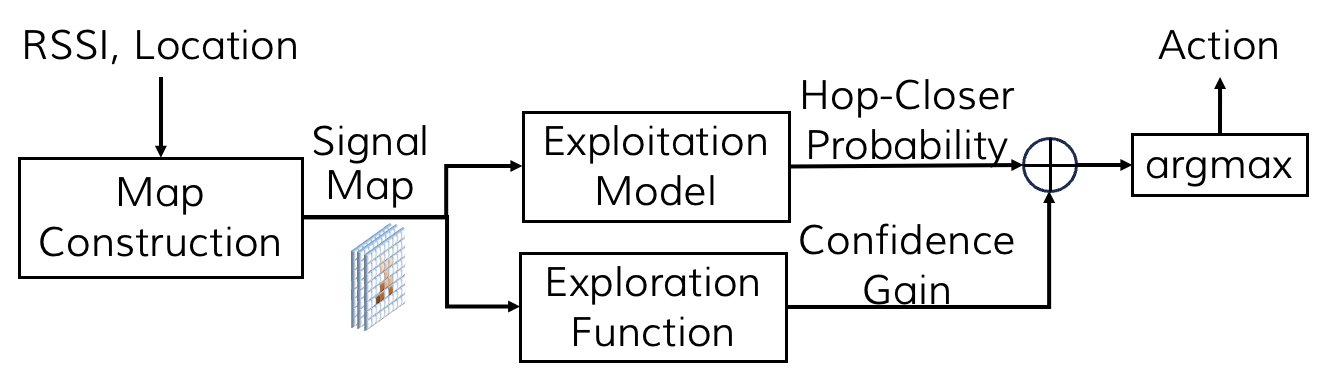}
  \caption{System overview of \sysname{}. }
  \label{fig:overview} 
\end{figure} 

We propose {\em \sysname{}}, a novel, deployable, zero-shot generalizable (ZSG) reinforcement learning model that leverages the RSSI spatial feature to 
enable robust and efficient search for a LoRa tag under RSSI fluctuation and domain shift. 
\FigureWord~\ref{fig:overview} illustrates the system diagram of \sysname{}. 
In each step, \sysname{} reconstructs the memory of RSSI measurements and agent locations into a map representation, termed a signal map. 
Based on the map, it chooses the action that maximizes the probability of the agent hopping toward a tag with increasing confidence. 
Our contributions are the tailored-made exploitation and exploration components, detailed as follows: 
\begin{itemize}[leftmargin=*] 
  \item {\em Robust exploitation model to maximize the probability of hopping toward a LoRa tag: }
  To generalize across environments, we  
  build a deep learning exploitation model with a large spatial receptive field that   
   extracts RSSI spatial variation and estimates the probability of actions leading the agent toward a tag. 
  The model is optimized by policy gradient, 
  along with a feature consistency loss function to account for robustness against RSSI fluctuations 
  and an imitation learning loss function to boost training efficiency. 

  \item {\em Effective exploration function to maximize the confidence gain of actions: } 
  The sample completeness of signal maps, or simply the map completeness, determines the confidence of exploitation, 
  where low confidence could cause loops in search path and result in large localization errors. 
  We propose an exploration function that quantifies the map completeness to estimate the confidence gain brought by an action, 
  inspired by the upper confidence bound (UCB)~\cite{auer2002finite}.   
  With this exploration, 
  each action is chosen by jointly maximizing the hop-closer probability and the confidence gain of exploitation.
  This avoids path looping, leading the agent toward the tag with increasing confidence.  
\end{itemize}

\begin{figure}[tp]
  \centering
  \includegraphics[width=0.9\linewidth]{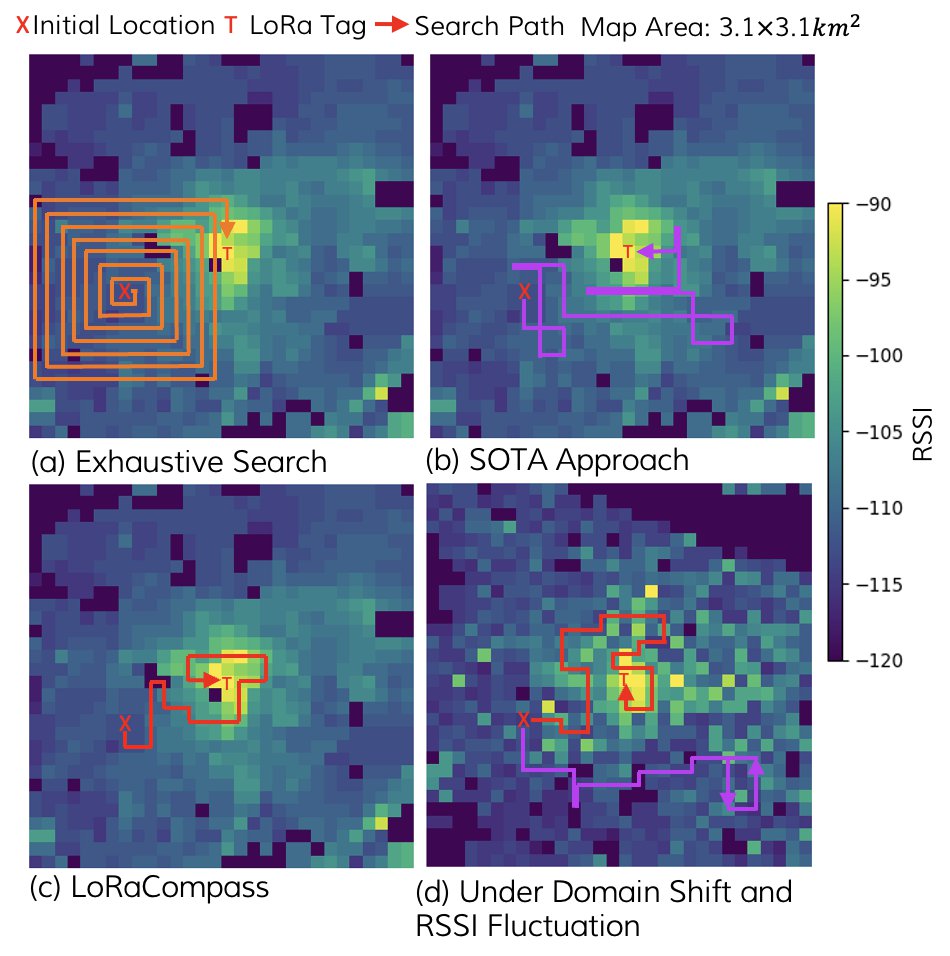}
  \caption{Search path from an agent at initial location $X$ to a LoRa tag at $T$, 
  with the RSSI heatmap shown as the background. 
  (a) Traditional exhaustive search, which is independent of RSSI and results in an inefficient path. 
  (b) State-of-the-art (SOTA) approach~\cite{soorki2024catch}, which achieves a shorter path under the same RSSI heatmap. 
  (c) Our \sysname{} achieves comparable path length for the same search case. 
  (d) Search path under domain shift and RSSI fluctuation. 
  The SOTA approach (purple line) fails to find the tag due to path looping, 
  whereas \sysname{} (red line) remains efficient and robust, successfully converging to the tag. } 
  \label{fig:region_coverage} 
\end{figure}

\FigureWord~\ref{fig:region_coverage} showcases the robustness and efficiency of \sysname{}, 
where a LoRa tag is placed in an urban region (labeled as $T$) with the underlying RSSI heatmap shown in the background. 
Starting from its initial location $X$, the agent must determine its next hop toward $T$, 
from five possible actions (\{N,E,S,W\} and stop), based on the RSSI measured at its current grid point. 
A traditional exhaustive search strategy is depicted in \FigureWord~\ref{fig:region_coverage}(a), 
where the agent follows a fixed “spiral” trajectory to cover the entire area. 
This approach is inefficient, incurring a search cost of $O(d^2)$ hops, where $d$ is the distance between the tag and the agent's initial location. 
In contrast, \FigureWord~\ref{fig:region_coverage}(b) shows a much shorter path produced by a SOTA approach, 
which optimizes each hop based on the RSSI sampled at the current grid point. 
While the SOTA approach is efficient in the given scenario, 
it lacks robustness and often fails to localize the tag under domain shift and RSSI fluctuation, as illustrated in \FigureWord~\ref{fig:region_coverage}(d) (purple line). 
By comparison, our \sysname{} remains robust and efficient in locating the LoRa tag, as shown in \FigureWord~\ref{fig:region_coverage}(c) and \FigureWord~\ref{fig:region_coverage}(d) (red line).

Empirically, we have validated \sysname{} with extensive experiments 
involving more than 20,000 search processes across both ground-based and drone-assisted scenarios 
at multiple diverse sites that covers more than $80\mbox{km}^2$ in total. 
\sysname{}, trained merely on any one site, achieves a high success rate in other unseen sites, 
successfully locating the LoRa tag within $100\mbox{m}$ over 90\% of the time,
outperforms SOTA approaches significantly by 40\%. 
Additionally, it demonstrates satisfactory efficiency, with the search path length grows linearly with the initial distance between the agent and the tag.  

The remainder of this paper is organized as follows. 
We define the problem in Section~\ref{sec:preliminary} 
and present \sysname{} design in Section~\ref{sec:method}, 
including system overview, map construction, exploitation model, and exploration function.
Then, we discuss experimental results in Section~\ref{sec:exp} and   
review related work in Section~\ref{sec:related_work}. Finally, we conclude in Section~\ref{sec:conclude}. 
In the Appendix, we provide more visualizations of the RSSI heatmaps used in experiments.

\section{Preliminaries}\label{sec:preliminary}
In this section, we present the problem formulation in \SectionWord~\ref{subsec:prob_overview} 
and discuss model training in \SectionWord~\ref{subsec:simulator}. 
The major symbols are summarized in \TableWord~\ref{table:notation}.

\begin{table}[tp]
  \caption{Summary of Major Symbols. }
  \footnotesize
  \renewcommand\arraystretch{1.1}
  \begin{tabular}{lp{6.4cm}}
    \hline
    \textbf{Notation}   & \textbf{Description} \\
    \hline
    $\mathcal{G}$                   & Global 2D coordinate system ($\mathcal{G}=\{(i,j)\mid i,j\in \mathbb{Z}\}$) \\
    $K$                             & Maximum number of steps in one search ($K\in \mathbb{Z}^+$) \\
    $k$                             & Index of search steps ($k=0,1,\ldots, K-1$) \\
    $c$                             & Grid coordinate of LoRa tag ($c\in\mathcal{G}$) \\
    $u_k$                           & Grid coordinate of agent at step $k$ ($u_k\in\mathcal{G}$) \\
    $s_k$                           & Vector of search state at step $k$ ($s_k=u_k-c$) \\
    $v_k$                           & RSSI measured at step $k$ ($v_k \in [-120, -70]\cap \mathbb{Z}$) \\
    $\overline{v}(u)$                & Mean of RSSI measurements at grid $u$ \\
    $a_k$                            & Action at step $k$ ($a_k\in \{\text{N},\text{E},\text{S},\text{W},\text{O}\}$) \\ 
    $r_k$                           & Reward at step $k$ ($r_k\in\{+1, -1\}$) \\
    \hline
  \end{tabular}
\label{table:notation}
\end{table}

\subsection{Problem Formulation}\label{subsec:prob_overview}
We index the grid points in a search area by a global 2D coordinate system, denoted as $\mathcal{G} = \{(i,j) \mid i,j \in \mathbb{Z}\}$, 
where the $i$-axis extends from west (W) to east (E), and the $j$-axis extends from south (S) to north (N). 
In a search process of $K$ steps, an agent (carrying a LoRa sensor) at an initial location $u_0 \in \mathcal{G}$ searches for a nearly immobile LoRa tag at $c \in \mathcal{G}$.  
In step $k=0,1,\ldots,K-1$, the agent at location $u_k \in \mathcal{G}$ obtains an RSSI measurement $v_k \in [-120, -70]\cap \mathbb{Z}$.\footnote{We set $v_k=-120$ for signal loss.} 
Then, it hops to the next location according to the action determined by a {\em decision-making process}: 
\begin{equation}
\{u_{0:k}, v_{0:k}\}\mapsto a_k,
\label{eq:policy_model}
\end{equation}
where $u_{0:k}, v_{0:k}$ are agent locations and the sequences of RSSI measurements, respectively, and $a_k\in\{\text{N}, \text{E}, \text{S}, \text{W}, \text{O}\}$, with ``O'' denoting ``stop''. 
In the optimal process, an agent always chooses the action that points to the tag (i.e., the tag's direction) and, upon arrival, takes ``O'' thereafter. 

\begin{figure}[tp]
  \centering
  \includegraphics[width=0.9\linewidth]{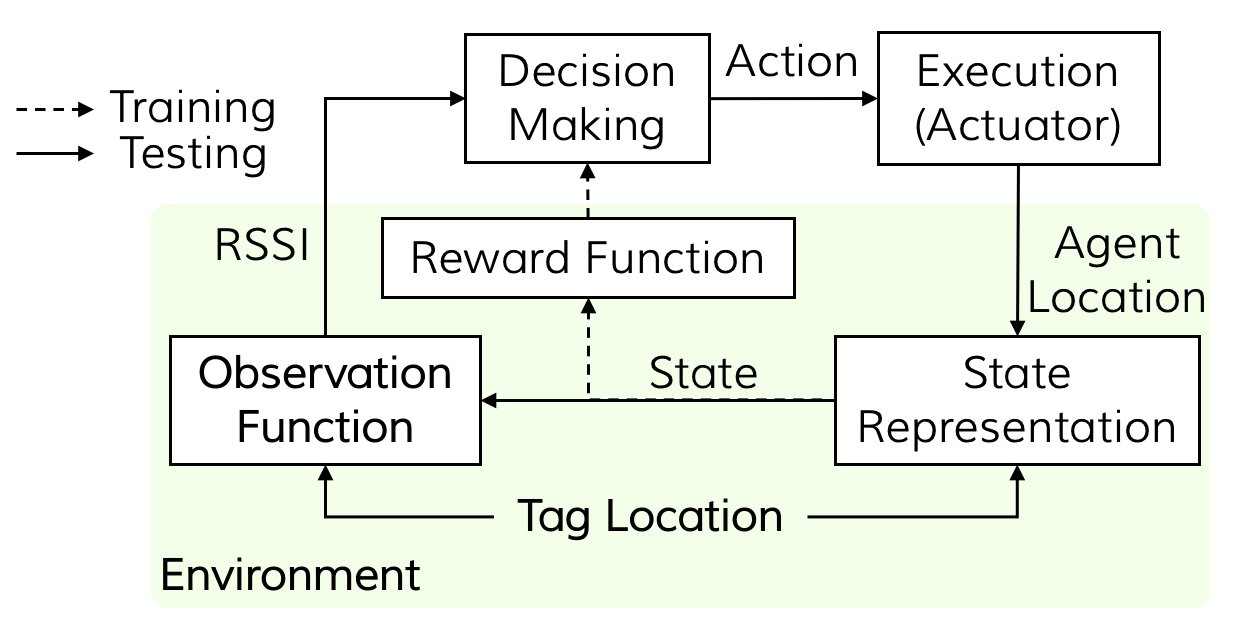}
  \caption{Illustration of problem formulation (CPOMDP). The decision-making must interact with four processes: action execution, state representation, observation function, and reward function. }
  \label{fig:prob_def} 
\end{figure} 

The decision-making process has to consider two key aspects: 
(1) it needs to use RSSI to infer the tag's direction, since the displacement from the agent to the tag is {\em not directly observable}; 
(2) it must account for the {\em context} of the search (i.e., environments), because each environment defines a unique mapping from the agent's location to RSSI. 
In light of these two aspects, we formulate this search problem as a contextual partially observable Markov decision process (CPOMDP)~\cite{tang2025deep}.   
The problem formulation is illustrated in \FigureWord~\ref{fig:prob_def}. 
Specifically, the agent must interact with four processes, detailed as follows:  
\begin{itemize}[leftmargin=*]
    \item {\em State representation: } 
    We represent the {\em state} of the search, denoted by~$s$, as the displacement vector between the agent ($u_k$) and tag ($c$): 
    \begin{equation}
        s_k=u_k-c.  
    \end{equation} 
    Once known, the state reveals the optimal action (i.e., the tag's direction).
    As mentioned, the states are not directly observable; however, we can measure RSSI from the state. 
    Thus, the search problem is {\em partially observable}.  
    \item {\em Observation function: } 
    In a state $s_k$, the agent receives a LoRa signal, from which it observes an RSSI measurement $v_k$.
    We refer to the process from $s_k$ to $v_k$ as the observation function. 
    Considering RSSI fluctuation and domain shift, 
    we model RSSI as a random variable sampled from a distribution that depends on the signal environment between the agent and the tag: 
    \begin{equation}
        v_k \sim P_c(v \mid s_k).   
        \label{eq:heatmap}
    \end{equation} 
    During one search (when the tag location $c$ is fixed), the distribution solely depends on the state $s_k$;
    in a different search, however, the distributions vary with the tag location $c$.  
    Therefore, we formulate the search for one tag location as a POMDP instance, 
    and the search problem for {\em any} tag location is a contextual POMDP (or CPOMDP), where the context is defined by $c$.

    \item {\em Action execution: } 
    After obtaining an RSSI measurement, the agent executes a chosen action ($a_k$) by hopping to the next location ($u_{k+1}$). 
    This execution process is expressed as:  
    \begin{equation}
        \{u_k, a_k\}\mapsto u_{k+1}. 
    \end{equation}
    In practice, 
    the agent may not faithfully execute the chosen action due to its subjective judgments or environmental restrictions (e.g., walls, highways, terrain). 
    To model this, we execute a chosen action with probability $\tau\in[0,1]$, 
    and otherwise draw uniformly from the action space ($\{\text{N}, \text{E}, \text{S}, \text{W}, \text{O}\}$). 
    This formulation validates the robustness of decision-making under actuation uncertainty. 
    The performance of our approach is validated in \FigureWord~\ref{fig:exp_sr_rand}. 

    \item {\em Reward function: } 
    After each decision, we give it a positive reward if it guides an agent toward a tag and penalize it otherwise. 
    The reward signal for step $k$ is: 
    \begin{equation}
        r_k=\left\{
        \begin{aligned}
            &+1, \quad &\lVert s_{k+1}\rVert < \lVert s_{k}\rVert\ \text{or}\ \lVert s_{k+1}\rVert=0,\\
            &-1, \quad &\text{otherwise}.  
        \end{aligned}
        \right.
        \label{eq:reward}
    \end{equation}
    Note that the reward signals are available for model training but is not used during testing. 
\end{itemize} 

Recall that our objective is to locate a LoRa tag in unknown environments in the fewest hops. 
We formulate this as maximizing the expected rewards in the search for any $c$: 
\begin{equation}
\max\ \mathbb{E}_{c \sim \mathcal{G}} \left[ \sum_{k=0}^{K-1} r_k \right],   
\label{eq:prob_def}
\end{equation}
where the total rewards of a search are within $[-K, K]$. 
To this end, we design a {\em zero-shot generalization (ZSG)} reinforcement learning model for decision-making in unknown environments. 
Specifically, with a {\em site} defined as the set of RSSI measurements and grid locations involved in the search for one tag location (i.e., one POMDP instance), 
our model, once trained on one site, should generalize to other unseen sites (maintaining performance comparable to the training site) without the need for fine-tuning. 

\subsection{Sim-to-Real Paradigm}\label{subsec:simulator} 

\begin{figure}[t]
    \centering
    \begin{minipage}{0.49\linewidth}
        \centering
        \includegraphics[width=\linewidth]{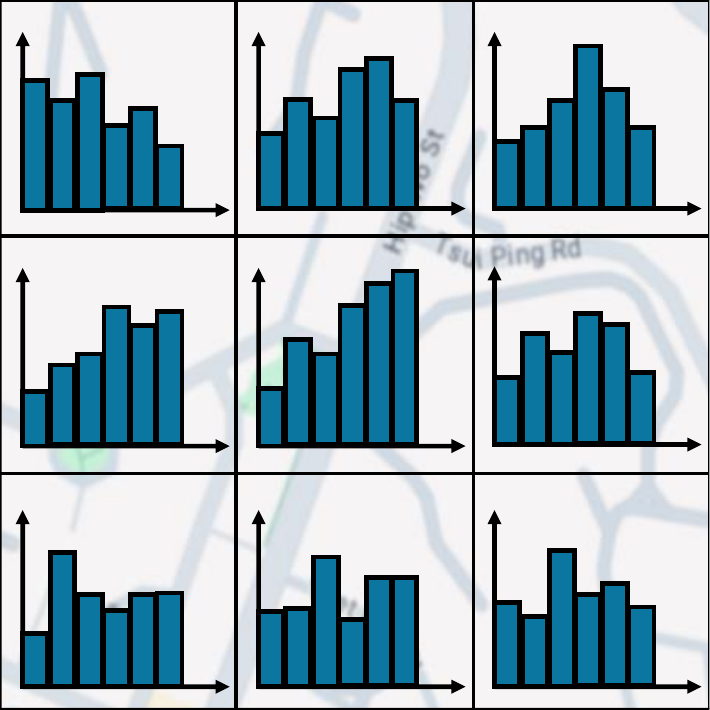}
        \caption{Illustration of the simulator for model training:  
        the observation function is simulated by RSSI histograms collected within grid points. }
        \label{fig:simulator} 
    \end{minipage}
    \hfill
    \begin{minipage}{0.49\linewidth}
        \centering
        \includegraphics[width=\linewidth]{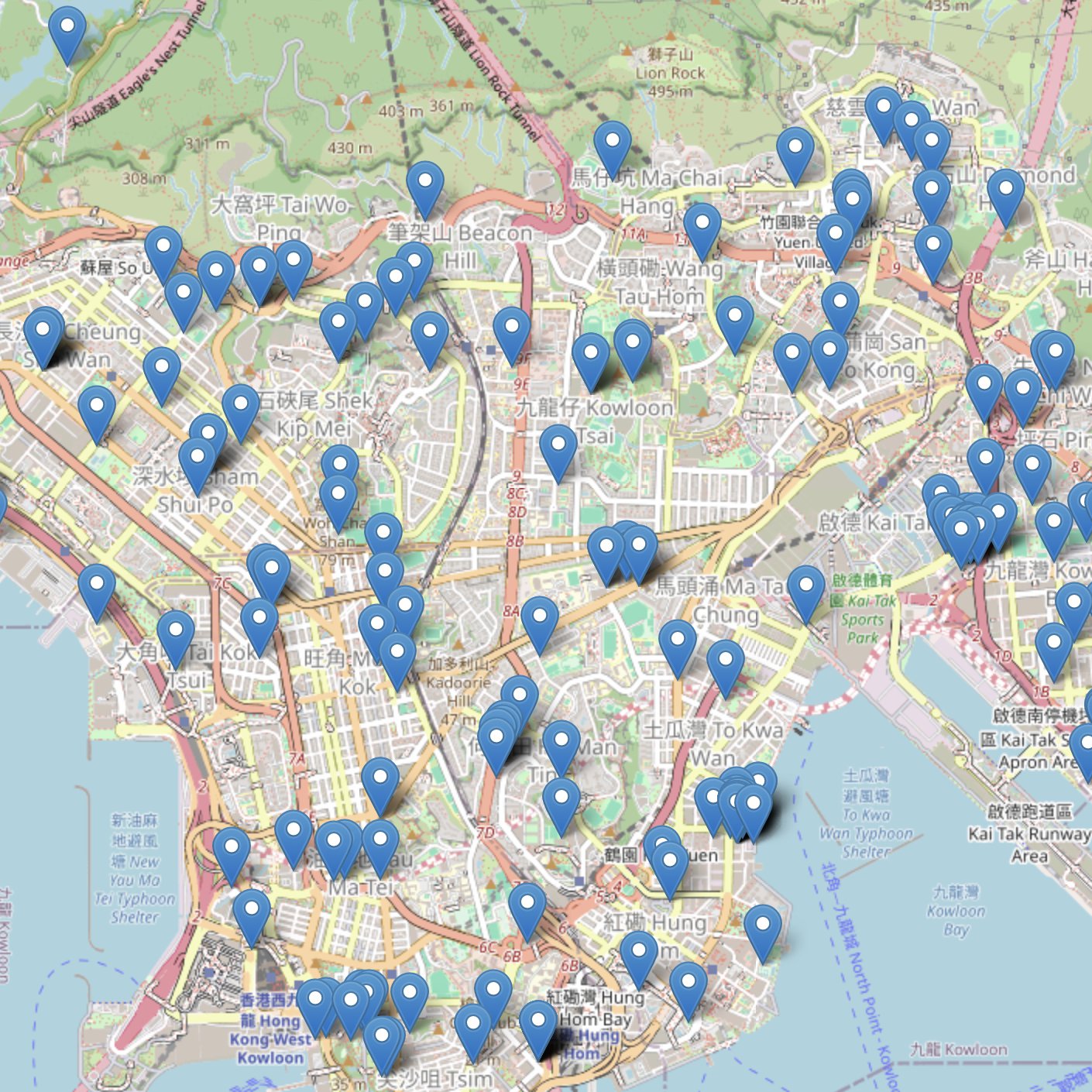}
        \caption{Distribution of LoRa gateway in an urban region of $40\mbox{km}^2$ (gateways are marked by blue labels). }
        \label{fig:gw_map}
    \end{minipage}
\end{figure}

Training a reinforcement learning model requires massive search processes, which is extremely time-consuming and costly in practice. 
To lower training costs, we
adopt a {\em sim-to-real paradigm}: the RL model is trained in a simulator and deployed in real-world testing. 
Thus, we build a simulator that implements the four processes described in \SectionWord~\ref{subsec:prob_overview},   
where the observation function is simulated with real-world data. 
During the data collection, we fix a LoRa tag and collect RSSI histograms at the tag's surrounding grid points, as shown in \FigureWord~\ref{fig:simulator}.  
During simulation, the simulated observation function generates RSSI by sampling from those histograms. 
In this way, our simulator is realistic and lightweight, sufficient for on-policy RL training. 

Furthermore, we can crowdsource the observation data to reduce the costs of building the simulator.  
\FigureWord~\ref{fig:gw_map} shows an example of LoRa gateway distribution in a city. 
Based on this infrastructure, we ask volunteers (e.g., caregivers, bus drivers, and postal workers) to wear LoRa tags and GPS devices 
and collect the RSSI histograms over an extended period (the observation function can be centered at either the sensor or the tag; see~\cite{aernouts2018sigfox} for more details). 
Last but not least, our model is context-efficient, using one training site to achieve satisfactory performance in unseen others (see \SectionWord~\ref{sec:exp}).

\section{\sysname{} Design}\label{sec:method}
In this section, we present the design of \sysname{}. First, we give an overview of \sysname{} in \SectionWord~\ref{subsec:method_overview}. 
Then, we discuss its map construction in \SectionWord~\ref{subsec:feature_extract}, exploitation model in \SectionWord~\ref{subsec:policy_model}, 
and exploration function in \SectionWord~\ref{subsec:explore}. 

\subsection{System Overview}\label{subsec:method_overview}

\FigureWord~\ref{fig:overview} illustrates the decision process of \sysname{}. 
In each step (indexed by $k$), it outputs an action ($a_k$)
using the memory of search data ($\{u_{0:k}, v_{0:k}\}$). 
First, it reconstructs the search data into a signal map, denoted by $\mathcal{M}_k$: 
\begin{equation}
    \{u_{0:k},v_{0:k}\}\mapsto \mathcal{M}_k.  
\end{equation}
Based on the map, it uses an exploitation model, denoted by $\pi_\theta$, to estimate the probability of an action ($a$) that leads the agent toward a tag (or simply the {\em hop-closer probability}), given as  
\begin{equation}
    \pi_\theta(a\mid \mathcal{M}_k),   
\end{equation}
where $\theta$ is the model parameter. In parallel, 
it employs an exploration function, denoted by $U$, to compute the confidence gain brought by an action:  
\begin{equation}
    U(a, \mathcal{M}_k). 
\end{equation}
Finally, it makes decisions by jointly maximizing the hop-closer probability and the confidence gain, expressed as  
\begin{equation}
    a_k = \operatorname*{arg\,max}_{a \in \{\text{N}, \text{E}, \text{S}, \text{W}, \text{O}\}} \left[ \pi_\theta(a \mid \mathcal{M}_k) + U(a, \mathcal{M}_k) \right]. 
\label{eq:final_decision}
\end{equation}
The following subsections will present the three computing blocks in detail. 

\subsection{Map Construction}\label{subsec:feature_extract}

\begin{figure}[t]
  \centering
  \includegraphics[width=0.7\linewidth]{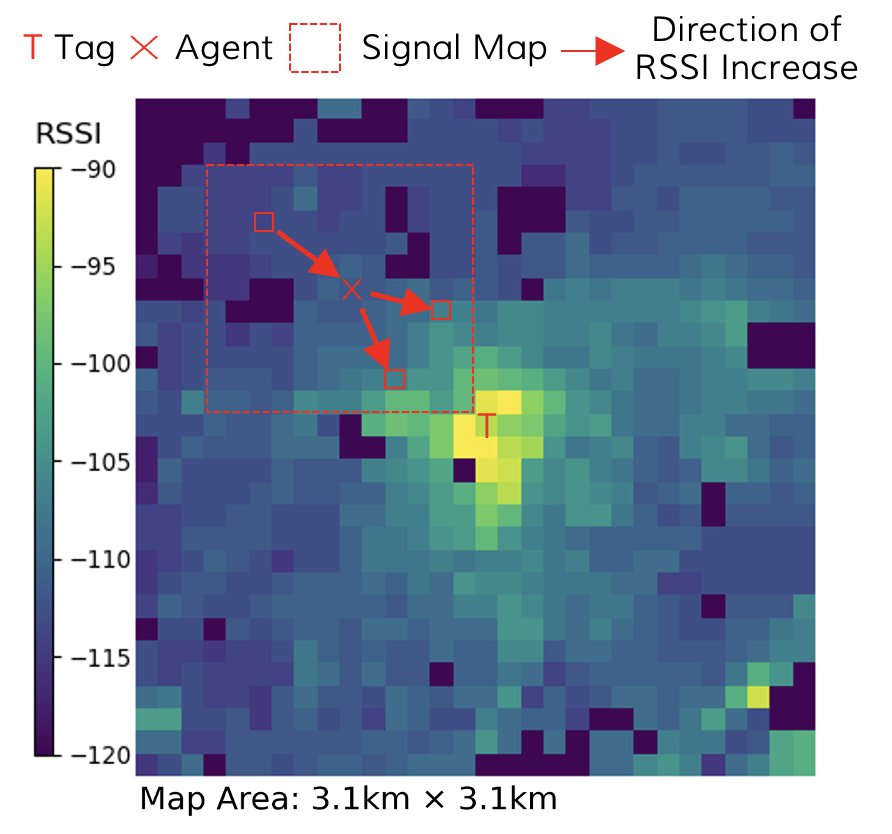}
  \caption{Illustration of signal map. }
  \label{fig:demo_feature} 
\end{figure}

We reconstruct the memory of search data ($\{u_{0:k}, v_{0:k}\}$) into a map representation, termed the signal map ($\mathcal{M}_k$). 
This intuition is illustrated in \FigureWord~\ref{fig:demo_feature}. 
Despite RSSI fluctuation and domain shift, RSSI in a large area shows a clear pattern: 
it is statistically higher when closer to a tag, and lower when further away from it. 
This pattern sheds light on how to determine the tag directions across environments, because it is based on 
the physical principle of signal attenuation, whereby signal strength decays over distance due to energy dissipation. 
In this way, the map representation embeds a generalizable inductive bias into the reinforcement learning model to robustly operate across environments. 

Formally, the map construction converts the search data into an {\em RSSI map} and a {\em sample map}, denoted as~$\mathcal{M}_k^{(v)}$ and~$\mathcal{M}_k^{(n)}$, respectively. 
The signal map is their concatenation:  
\begin{equation}
    \mathcal{M}_{k}=\left[\mathcal{M}_k^{(v)}, \mathcal{M}_k^{(n)}\right]. 
    \label{eq:map_concate}
\end{equation} 

The RSSI map reflects the RSSI spatial distribution based on the search data.  
Let $v(u)$ denote an RSSI measurement sampled at grid $u$, 
and $\overline{v}(u)$ denote the mean of all RSSI samples at $u$ (i.e., for all $u$ in $u_{0:k}$). 
Rewriting agent location as $u_k=(i_k, j_k)$, the RSSI map is given as  
\begin{equation}
    \mathcal{M}_k^{(v)}=\left\{\overline{v}(i,j)\ \big|\ \left|i-i_k\right|, \left|j-j_k\right|< m\right\}, 
    \label{eq:signal_map}
\end{equation}
where $m\in\mathbb{Z}^+$ controls the map size (see parameter study in \FigureWord~\ref{fig:exp_param_study_perc}). 
Here, the RSSI at unvisited grid points is set to the minimum value. 

The sample map reflects the sample completeness of the RSSI map.   
Let $\text{Num}(u, U)$ denote the number of times element $u$ appears in a sequence $U$. 
The sample map is given as 
\begin{equation}
\begin{aligned}
        \mathcal{M}_k^{(n)}
        =\left\{\text{Num}\left((i,j), u_{0:k}\right)\ \big|\ \left|i-i_k\right|, \left|j-j_k\right|< m\right\}.  
\end{aligned}
\end{equation}

\subsection{Exploitation Model}\label{subsec:policy_model} 

\begin{figure}[tp]
  \centering
  \includegraphics[width=0.9\linewidth]{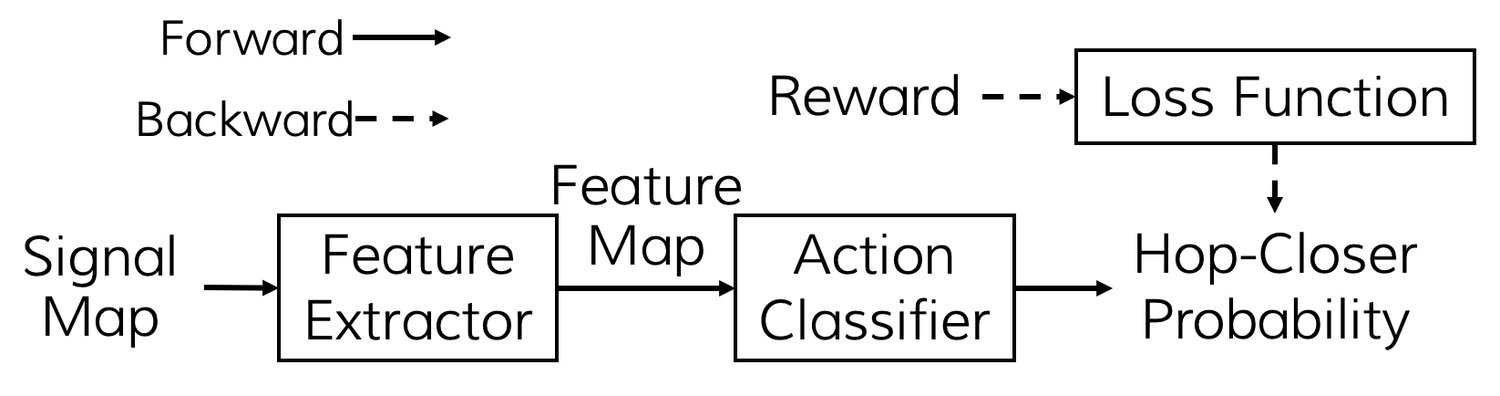}
  \caption{System diagram of the exploitation model. }
  \label{fig:exploit_model} 
\end{figure}

Given a signal map, the exploitation model aims to estimate the hop-closer probability $\pi_\theta(a\mid \mathcal{M}_k)$. 
The system diagram is shown in \FigureWord~\ref{fig:exploit_model}.   
In the forward path, the model first extracts a feature map, denoted as $\widetilde{\mathcal{M}}_k$; then, it 
employs a deep learning action classifier to estimate the hop-closer probability. 
 In the backward path, a loss function uses reward signals ($r_k$) to supervise the learning of the action classifier. 
 Below, we present the details of feature extraction, action classification, and loss function. \\

\noindent
{\bf Feature Extraction: }
We preprocess the signal map into a feature map to improve the generalization ability of the action classification. 
First, given the RSSI map, we highlight the RSSI spatial variation relative to the agent's current location, which is defined as
\begin{equation} 
    \widetilde{\mathcal{M}}_k^{(v)}=\left\{v-\overline{v}{(u_k)}\ \big|\ v\in\mathcal{M}_k^{(v)}\right\}.
\end{equation} 
With this feature, the agent is more likely to search for the areas whose RSSI is larger than its current location. Yet, it also effectively prevents 
the action classifier from focusing on the shortcuts from RSSI values to tag directions. 

Second, we remove the information of action sequence from the sample map. 
Specifically, the number of visits on the sample map captures the sequence of previous actions. 
Once the sequence is considered as the feature of tag directions, the agent may prefer to repeat its search trajectories in training, while overlooking current RSSI measurements.  
This degrades the generalization outside training.  
To address this, we eliminate the sequential information by binarizing the number of visits into visited and unvisited states:  
\begin{equation}
\begin{aligned}
        \mathcal{M}_k^{(b)}
        =\left\{\mathbbm{1}\left((i,j)\in u_{0:k}\right)\ \big|\ \left|i-i_k\right|, \left|j-j_k\right|< m\right\},   
\end{aligned}
\end{equation}
where $\mathbbm{1}(\cdot)$ returns $1$ if the criterion is met and $0$ otherwise. 

Empirically, we concatenate the these features into a feature map:  
\begin{equation}
    \widetilde{\mathcal{M}}_{k}=\left[\mathcal{M}_k^{(v)}, \widetilde{\mathcal{M}}_k^{(v)}, \mathcal{M}_k^{(b)}\right]. 
    \label{eq:feature_map}
\end{equation} 
Our ablation study shows that this feature map improves the generalization ability, as it 
highlights the RSSI spatial variation and removes the sequential information of actions, 
without causing information loss (see \FigureWord~\ref{fig:exp_ablation_study_map}). \\ 

\noindent
{\bf Action Classification: } 
To leverage both high-level and low-level spatial patterns in the feature map, we design a deep learning action classifier to estimate the hop-closer probability.
First, the classifier uses a convolutional neural network (CNN), denoted by $\pi_{\theta_1}$, to aggregate the feature map into a search feature, given as 
\begin{equation}
    z_k=\pi_{\theta_1}(\mathcal{M}_k),
    \label{eq:search_feature}
\end{equation}
where $z_k$ is the search feature at step $k$. Then, it employs a multilayer perceptron (MLP) 
with a Softmax function, denoted by $\pi_{\theta_2}$, which maps the search feature to the hop-closer probability, shown as
\begin{equation}
    \pi_\theta(a\mid\mathcal{M}_k)=\pi_{\theta_2}(a\mid z_k), 
\end{equation}
where $\theta=[\theta_1, \theta_2]$. Note that the probabilities are normalized over action space. 

In our implementation, the CNN has three layers, each using $3\times 3$ kernels, with both padding and stride set to $1$. 
The number of channels are $16$, $32$ and $64$, respectively, and the MLP has three layers with a feature size of $128$. 
Here, our design follows standard practice for processing map data;   
other model structures and hyperparameters may also be applicable~\cite{kirk2023survey, guo2024dynamic, zhao2024review}. \\

\noindent
{\bf Loss Function: } As mentioned, we train the action classifier via on-policy learning, facilitated by the simulator introduced in \SectionWord~\ref{subsec:simulator}. 
Recall that $r_k$ is the reward signal at step $k$. The policy gradient (PG) loss of a search process is calculated as\footnote{Here, the reward signal directly define the advantage function.}
\begin{equation}
\mathcal{L}_\text{PG}(\theta) = -\sum_{k=0}^{K-1} r_k \log \pi_\theta(a_k \mid \mathcal{M}_k).     
\label{eq:pg_loss}
\end{equation}
To account for the RSSI fluctuation and training efficiency, 
we additionally design a feature-consistency loss function and an imitation-learning loss function. 

\begin{figure}[t]
  \centering
  \includegraphics[width=0.6\linewidth]{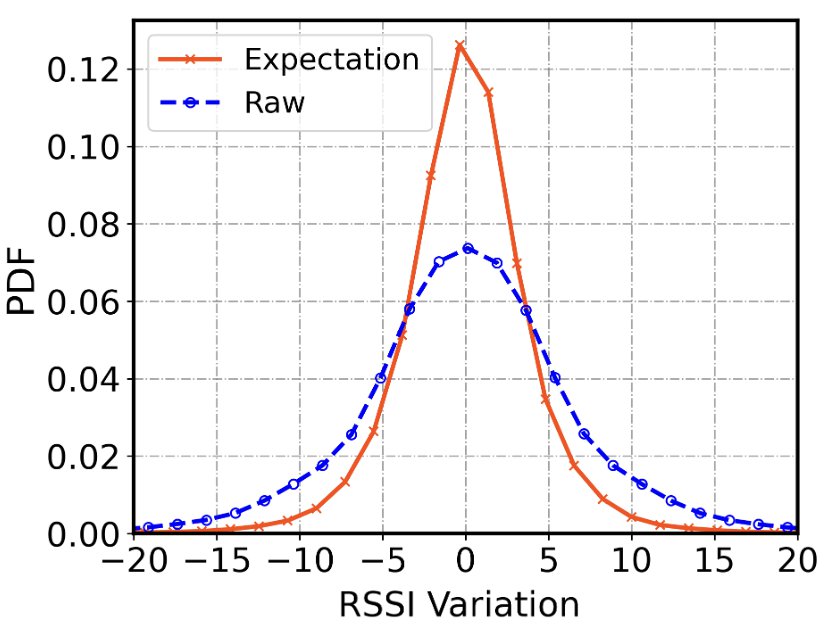}
  \caption{RSSI variation between straight-neighboring grids in noise-based and noise-free environments. }
  \label{fig:demo_pd} 
\end{figure}

First, trained with only policy gradient, the action classifier is vulnerable to RSSI fluctuation, resulting in degraded search efficiency. 
Specifically, since RSSI fluctuation perturbs the RSSI spatial pattern, 
the agent may have to sample a grid point multiple times, leading to inefficient search. 
To address this, we make the search feature ($z_k$) robust against RSSI fluctuations. 
Our idea is simple: for each real, noise-based environment, we prepare a simulated, noise-free environment.   
By running the agent in both environments, we minimize the discrepancy of the extracted search features. 
This is implemented as a feature-consistency (FC) loss function, shown as 
\begin{equation} 
    \mathcal{L}_\text{FC}(\theta_1)=\sum_{k=0}^{K-1}\mathrm{MSE}\left(z_k,z_k'\right), 
    \label{eq:SD_loss}
\end{equation}
where $z_k'$ is the search feature in the noise-free environment, and $\mathrm{MSE}$ calculates the mean squared error.  
Fundamentally, this loss function encourages the action classifier to predict noise-free RSSI ($\overline{v}$) 
using the noisy measurements ($v$) across grid points. 
This reduces the gap that    
RSSI variations in noise-based environments are expected to be larger than those in the noise-free counterpart, as shown in \FigureWord~\ref{fig:demo_pd}.  

Second, we use imitation learning to accelerate the training of the action classifier. 
Specifically, for RL training, we have two choices: 
policy-gradient training to learn through trial-and-error, and imitation learning to learn from the optimal actions. 
In comparison, the policy-gradient training is often slow, yet  
imitation learning does not capture the stochastic nature of the search~\cite{le2022deep}. 
To combine their merits, we adopt a trick: 
we train the model using policy gradient for better performance and employ imitation learning (IL) as an auxiliary supervision to accelerate model training.  
Recall that $s_k$ is the search state that reveals the optimal actions, 
we implement the auxiliary IL loss function in feature space, shown as  
 \begin{equation}
    \mathcal{L}_\text{IL}(\theta_1)=\sum_{k=0}^{K-1}\sum_{a}\left| Pr(a\mid z_k)-Pr(a\mid s_k)\right|.   
    \label{eq:SL}
\end{equation}
To facilitate the imitation, we add one fully connected layer to compute $Pr(a\mid z_k)$. 

Overall, the loss function of action classification is 
\begin{equation}
    \mathcal{L}(\theta)=\mathcal{L}_\text{PG}(\theta)+\omega_1\mathcal{L}_\text{FC}(\theta_1)+\omega_2\mathcal{L}_\text{IL}(\theta_1).  
    \label{eq:overall_loss}
\end{equation} 
In the function, $\omega_1,\omega_2$ are the weights for feature consistency loss and imitation learning loss; 
their parameter studies are shown in~\FigureWords~\ref{fig:exp_param_study_pd} and~\ref{fig:exp_training_eff}, respectively.  

\subsection{Exploration Function}\label{subsec:explore}

\begin{figure}[t]
  \centering
  \includegraphics[width=\linewidth]{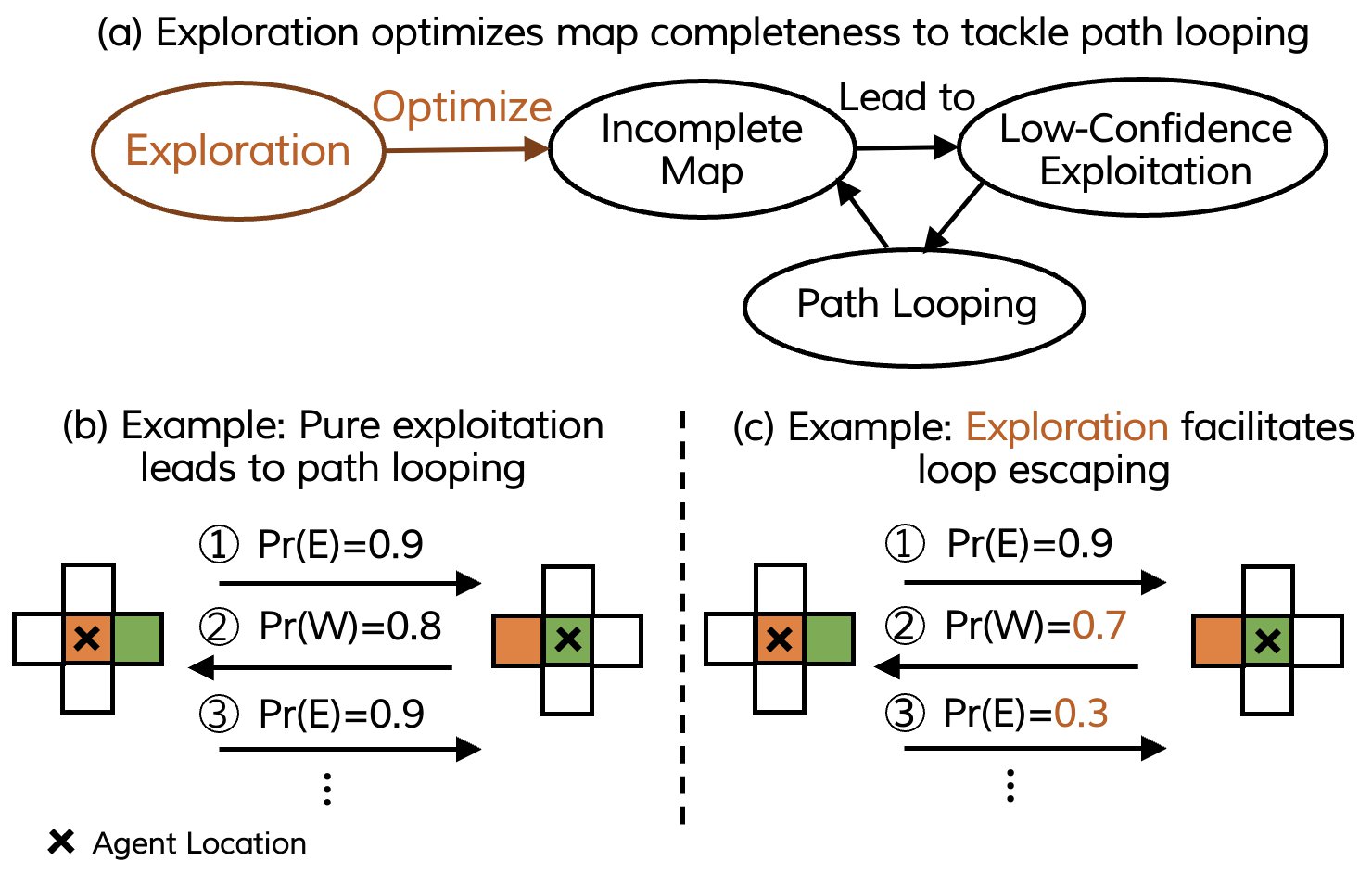}
  \caption{(a) illustrates the issue of path looping (Incomplete signal map leads to low-confidence exploitation that may form path looping. 
    In turn, path looping stagnates in completing the signal map, making it difficult to escape. 
    To escape loops, we introduce an exploration function to optimize the map completeness). We exemplify path looping and loop escaping in (b) and (c), where  
    probabilities are normalized across exploitation and exploration components. }
  \label{fig:demo_overfit} 
\end{figure}

The exploitation relies on the completeness of signal map. 
If executed alone (i.e., $a_k=\arg\max_a \pi_\theta(a\mid \mathcal{M}_k)$),  
it often suffers the issue of {\em path looping}. As illustrated in \FigureWord~\ref{fig:demo_overfit}(a),       
early in the search, since the signal map is highly incomplete, 
the exploitation has to classify actions with low confidence, and the resulting random actions may form loops in the search path. 
Once trapped in a loop, the agent stagnates in completing the signal map, which, in turn, makes it harder to escape from the loop, as illustrated in \FigureWord~\ref{fig:demo_overfit}(b). 
The path looping could lead to premature convergence and inefficiencies, culminating in   
large localization errors and overall search failures. 

To escape loops, we design an exploration mechanism to optimize the {\em map completeness}. 
With the exploration, each action is chosen by jointly maximizing the hop-closer probability for exploitation 
and the map completeness for exploration (i.e., $a_k=\arg\max_a \left[\pi_\theta(a\mid \mathcal{M}_k)+U(a,\mathcal{M}_k)\right]$). 
This drives an agent to leave a loop to complete the map, 
thereby addressing the path looping issue, as shown in \FigureWord~\ref{fig:demo_overfit}(c).   
To implement this, we design an exploration function (i.e., $U(a, \mathcal{M}_k)$) to evaluate the map completeness. \\

\noindent
{\bf Map Completeness: }
As mentioned, an incomplete map leads to low confidence in exploitation, resulting in path looping. 
To address the issue, we evaluate map completeness in terms of {\em exploitation confidence}. 
Specifically, the exploitation confidence depends on both the aleatoric uncertainty (which arises from the mapping from {\em complete} signal maps, denoted as $\mathbb{E}(\mathcal{M}_k)$, to actions):
\begin{equation}
    \mathbb{E}(\mathcal{M}_k)\mapsto a,  
    \label{eq:aleatoric}
\end{equation}
and epistemic uncertainty (which reflects the discrepancy in exploitation given incomplete and complete signal maps):\footnote{Aleatoric uncertainty is inherent in a system, and epistemic uncertainty is due to lacking certain knowledge~\cite{jiang2022uncertainty}. } 
\begin{equation}
    \sum_a \left| \pi_\theta\left(a\mid \mathcal{M}_k \right)-\pi_\theta\left( a\mid \mathbb{E}(\mathcal{M}_k) \right) \right|.     
    \label{eq:exploit_uncertainty}
\end{equation}
The aleatoric uncertainty can be largely reduced, so long as we choose a large map, a small action space, and adequate model capacity for the action classifier. 
Therefore, to improve exploitation confidence, 
the agent needs to minimize the main cause of the epistemic uncertainty, i.e., the {\em map incompleteness}, defined as  
the discrepancy of a RSSI map ($\mathcal{M}_k^{(v)}=\{\overline{v}\}$) to its complete counterpart ($\mathbb{E}(\mathbb{\mathcal{M}}_k^{(v)})=\{\mathbb{E}(v)\}$):    
\begin{equation}
\sum_{\overline{v}\in \mathcal{M}_k^{(v)}} \left| \overline{v}-\mathbb{E}(v) \right|,      
\label{eq:incompleteness}
\end{equation}
note that $\mathbb{E}(v)=\mathbb{E}(\overline{v})$. 
Here, we do not include the sample map since it merely assists in exploitation (also, a complete map has infinite samples, making direct calculation infeasible). 

However, we cannot directly compute the map incompleteness, 
since we lack complete maps. To address this, 
we estimate map incompleteness using the {\em number of visits} recorded by the sample map (i.e., $n\in \mathcal{M}_k^{(n)}$),  
because the visit counts determine the extent to which mean RSSI~($\overline{v}$) approximates the RSSI expectation~($\mathbb{E}(v)$). 
Formally, for each grid $u$, we estimate the upper bound of the grid-wise map incompleteness, denoted as $\epsilon(u)$, using
Hoeffding's inequality~\cite{auer2002finite}:  
\begin{equation}
\begin{aligned}
    Pr&\left(\left|\overline{v}(u)-\mathbb{E}\left[v(u)\right]\right|\ge\epsilon(u)\right)\\
    &\leq 2\exp{\left[-\frac{2\epsilon^2(u)n(u)}{v_{range}^2}\right]},  
\end{aligned}
\end{equation}
where $\overline{v}(u)$ and $n(u)$ are the mean RSSI and visit count at grid $u$, respectively, 
and the dynamic range $v_{range}=50$ in our case.  
With a confidence level of $\sigma\in [0,1]$, the estimated map incompleteness at $u$ is given as 
\begin{equation}
    \epsilon(u)=v_{range}\sqrt{\frac{-\log(\sigma/2)}{2n(u)}}. 
\label{eq:rssi_discrepancy}
\end{equation} 

Given \EquationWords~(\ref{eq:incompleteness}) and~(\ref{eq:rssi_discrepancy}), 
we design an exploration function to estimate 
the {\em reduction of map incompleteness} brought by an action~$a$ based on the current map $\mathcal{M}_k$. 
Let $n_a$ represent the number of visits at the next agent location ($u_{k+1}$) when action $a$ is executed.\footnote{The definition is properly adjusted to avoid the issue of division by zero. }   
The exploration function is given as 
\begin{equation}
\begin{aligned}
    U&(a, \mathcal{M}_k)= \sum_{\overline{v}\in \mathcal{M}_k^{(v)}} \left| \overline{v}-\mathbb{E}(v) \right|-\sum_{\overline{v}\in \mathcal{M}_{k\mathbf{+1}}^{(v)}} \left| \overline{v}-\mathbb{E}(v) \right|  \\
    &\approx v_{range}\sqrt{\frac{-\log(\sigma/2)}{2}}\left( \sqrt{\frac{1}{n_a}}-\sqrt{\frac{1}{n_a+1}} \right).     
\end{aligned}
\label{eq:explore_func_1}
\end{equation}
Fundamentally, since map incompleteness is the main source of exploitation confidence, 
the reduction in map incompleteness can be used to estimate the {\em confidence gain} of exploitation. \\

\begin{figure}[t]
  \centering
  \includegraphics[width=0.8\linewidth]{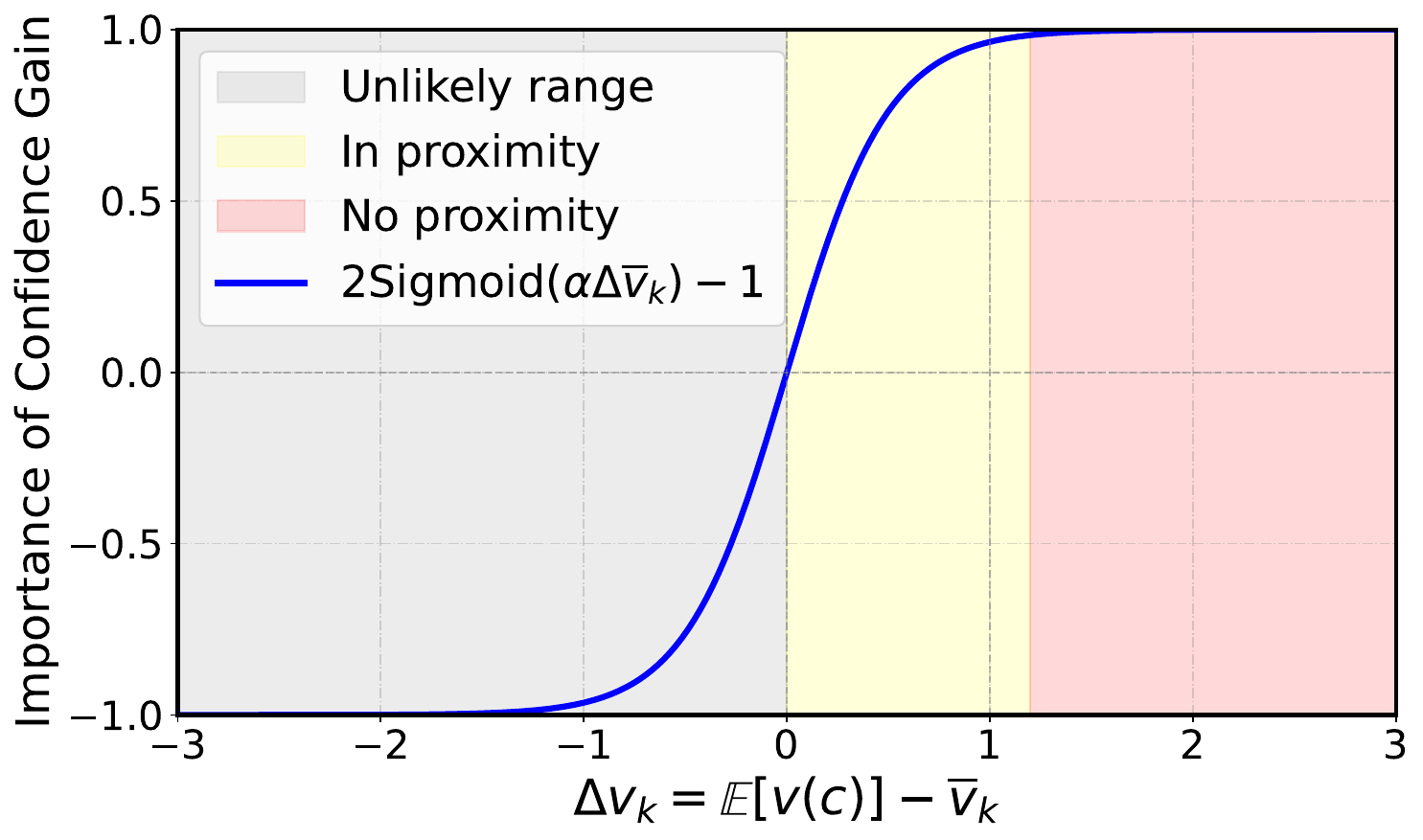}
  \caption{Illustration of using RSSI to tune the importance of confidence gain:  
  when RSSI is close enough to the reference value, we reduce the importance of confidence gain. }
  \label{fig:demo_confidence} 
\end{figure}

\noindent
{\bf Confidence Level: } 
In \EquationWord~(\ref{eq:explore_func_1}), the term involving the variation in visit counts, i.e., $\sqrt{1/n_a}-\sqrt{1/(n_a+1)}$, determines the confidence gain due to an action $a$, 
while the confidence level $\sigma$ determines the {\em importance of confidence gain}, i.e., $v_{range}\sqrt{-\log(\sigma/2)/2}$.  
Intuitively, when an agent is far away from a tag, we need to increase the importance of confidence gain to break path looping;   
however, when the agent is in proximity to a tag, we should decrease the importance, such that the search can quickly converge. 
Hence, we tune the importance of confidence gain according to agent's {\em proximity state} to the tag. 

\begin{figure*}[tp]
  \centering
    \centering
    \includegraphics[width=0.85\linewidth]{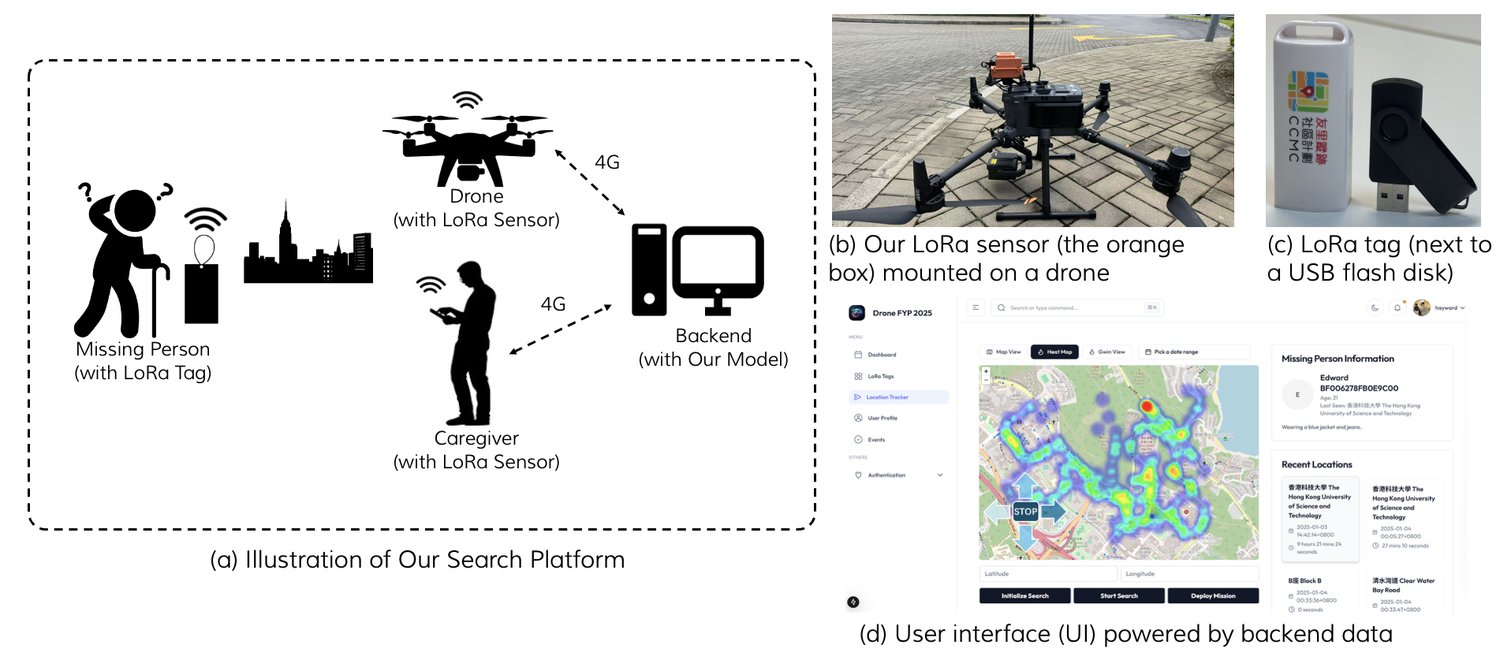}
    \caption{Our system deployment is illustrated in (a). 
    In the drone search scenario, a LoRa sensor is mounted on a drone (b) to detect signals from a lightweight IoT tag (c), worn by a missing individual. 
    Upon receiving a signal, the sensor reports its RSSI and GPS location to the backend system via 4G networks 
  (d, showing dummy personal data). The backend server, integrated with our \sysname{}, 
  guides the drone or ground agents on their next hops. }
    \label{fig:demo_deploy}
\end{figure*}

In our observation, when an agent is in proximity to a tag, it obtains high RSSI values with a high chance (see examples in Appendix).  
This allows us to capture the proximity state using RSSI values.
Recall that $\overline{v}_k$ is the mean RSSI at the current agent location, and let $\mathbb{E}[v(c)]$ represent 
the RSSI expectation at the tag location (this can be empirically obtained as a constant).  
The proximity state can be reflected by the gap between them, shown as 
\begin{equation}
    \Delta v_k=\mathbb{E}[v(c)]-\overline{v}_k. 
\end{equation} 
Since RSSI is noisy, we determine the proximity state by a soft threshold on $\Delta v_k$ based on the Sigmoid function.\footnote{We empirically adopt the Sigmoid function; other soft thresholds may also work.}
As shown in \FigureWord~\ref{fig:demo_confidence}, we assign higher importance to large $\Delta v_k$ (pink region) and 
lower importance for sufficiently small $\Delta v_k$ (yellow region). 
To enforce this, we rewrite the confidence level ($\sigma$) as a function of $\Delta v_k$, shown as  
\begin{equation}
    \sigma(\Delta v_k)=\exp\left[-\left(\text{Sigmoid}(\alpha \Delta v_k)-\frac12\right)^2\right],   
    \label{eq:sigma_function}
\end{equation}
where $\alpha\in \mathbb{R}$ controls the soft threshold (see parameter study in \FigureWord~\ref{fig:exp_param_study_alpha}). 

Applying \EquationWord~(\ref{eq:sigma_function}) to \EquationWord~(\ref{eq:explore_func_1}), 
the {\em exploration function} is given as 
\begin{equation}
        U(a, \mathcal{M}_k)=\beta\frac{e^{\alpha\Delta v_k}-1}{e^{\alpha\Delta v_k}+1}\left(\frac{1}{\sqrt{n_a}}-\frac{1}{\sqrt{n_a+1}}\right), 
\label{eq:explore_func}
\end{equation}
where $\beta\in\mathbb{R}$ is a weight for exploration (see parameter study in \FigureWord~\ref{fig:exp_param_study_ucbw}).   
Our exploration function addresses path looping without compromising search efficiency, which is 
validated in \FigureWord~\ref{fig:exp_comp_explore}. 

\section{Illustrative Experimental Result}\label{sec:exp}

In this section, we introduce our experimental results to validate \sysname{}. 
We introduce the experimental setting in \SectionWord~\ref{subsec:exp_set}, 
present system performance in \SectionWord~\ref{subsec:exp_sim}, 
and discuss the parameter study in \SectionWord~\ref{subsec:exp_param}. 

\subsection{Experimental Setting}\label{subsec:exp_set}

We implement \sysname{} through the sim-to-real paradigm, as mentioned eariler.  
First, we train and test \sysname{} in the real-data simulator in \SectionWord~\ref{subsec:simulator}, based on which 
we show parameter studies and validate its generalization ability to unseen sites through extensive quantitative results.  
Then, we deploy the trained \sysname{} to an existing system that consists of ground-based and drone-assisted search modes, 
so as to demonstrate its practicability in real-world deployment. 

The real-data experiments are based on five $4\times 4\,\mbox{km}^2$ sites located in different areas of a city, each covering diverse environments
including buildings, parks, streets, vegetation, and other landscapes~\cite{aernouts2018sigfox}. 
The five sites contain $345{,}700$ RSSI measurements crowdsourced by vehicles, whose heatmaps are visualized in Appendix~\ref{sec:appendix_heatmap}. 
In the experiment, we train \sysname{} on one site and test it on others, such that all testing sites are unseen to the model. 
In each site, 
the experiment involves multiple search processes, in which  
the agent starts from a random location whose {\em initial distance} to the tag (i.e., $\left\lVert s_0\right\rVert_2$) is uniformly distributed from $200\,\mbox{m}$ to $2{,}500\,\mbox{m}$. 
In total, our testing results are aggregated from more than $20{,}000$ search processes in the five sites. 

Our real-world deployment is illustrated in \FigureWord~\ref{fig:demo_deploy}(a). 
In the experiment, a target individual carries a lightweight LoRa tag (\FigureWord~\ref{fig:demo_deploy}c), 
which emits signals approximately once per minute. When the target is reported missing, the alert will be disseminated through the platform, 
activating volunteers to search for the tag. In a drone-assisted search, 
a drone equipped with a LoRa sensor (\FigureWord~\ref{fig:demo_deploy}b) listens for tag signals. 
Upon receiving a signal, the sensor reports the RSSI and GPS location to our backend system 
via a 4G network (\FigureWord~\ref{fig:demo_deploy}d). The backend, powered by \sysname{}, guides the drone to the next move. 
The ground searches operate similarly, yet the LoRa sensor is carried by a caregiver. 
Our ground-search experiment is in a $4\times 4\,\mbox{km}^2$ residential area 
(including residential blocks, shopping malls, sidewalks, overpasses, and parks), and 
the drone-search experiment is on a university campus (with a legal flying permit). 

\begin{table*}[t]
    \centering
    \caption{Success rate of comparison schemes.}
    \label{table:sr_1}
    \begin{threeparttable}
    \begin{tabular}{@{}c|ccccc@{}}
        \toprule
         &  Ranging & Simplex Method & RM Method & CMIYC (DRL) & \textbf{\sysname{}} \\         
        \hline
         Site 1${}^*$ & 0.042 $\pm$ 0.010  & 0.235 $\pm$ 0.013 & 0.502 $\pm$ 0.018& 0.683 $\pm$ 0.018 & \textbf{0.996} $\pm$ 0.003  \\
        \hline
        \text{Site 2} & 0.022 $\pm$ 0.009  & 0.269 $\pm$ 0.037  & 0.477 $\pm$ 0.016  & 0.385 $\pm$ 0.021 & \textbf{0.910} $\pm$ 0.009 \\
        \text{Site 3} & 0.023 $\pm$ 0.005 & 0.222 $\pm$ 0.005 & 0.439 $\pm$ 0.026 & 0.366 $\pm$ 0.015& \textbf{0.974} $\pm$ 0.006 \\
        \text{Site 4} & 0.029 $\pm$ 0.005  & 0.314 $\pm$ 0.022 & 0.55 $\pm$ 0.022 & 0.375 $\pm$ 0.012  & \textbf{0.946} $\pm$ 0.015\\
        \text{Site 5} &  0.026 $\pm$ 0.006  & 0.224 $\pm$ 0.028 & 0.436 $\pm$ 0.020 & 0.368 $\pm$ 0.009 & \textbf{0.918} $\pm$ 0.016 \\
        \textbf{Mean} & 0.027 $\pm$ 0.002 & 0.253 $\pm$ 0.009 & 0.479 $\pm$ 0.013 & 0.373 $\pm$ 0.007 & \textbf{0.940} $\pm$ 0.007  \\
        \bottomrule
    \end{tabular}
    \begin{tablenotes}
        \small
        \item[*] Training Site
    \end{tablenotes}
    \end{threeparttable}
\end{table*}

\begin{table}[tp]
    \centering
    \caption{Success rate based on a different training site.}
    \label{table:sr_2}
    \begin{threeparttable}
        \begin{tabular}{@{}c|ccc@{}}
        \toprule
           & Ranging &  CMIYC (DRL) & \textbf{\sysname{}} \\         
        \hline
         Site 2${}^*$ & 0.030 $\pm$ 0.014  & 0.654 $\pm$ 0.002 & \textbf{0.981} $\pm$ 0.006  \\
        \hline
        \text{Site 1} & 0.016 $\pm$ 0.006  & 0.322 $\pm$ 0.008 & \textbf{0.936} $\pm$ 0.006 \\
        \text{Site 3} & 0.029 $\pm$ 0.007  & 0.346 $\pm$ 0.012& \textbf{0.960} $\pm$ 0.010 \\
        \text{Site 4} & 0.029 $\pm$ 0.009  & 0.289 $\pm$ 0.026  & \textbf{0.923} $\pm$ 0.010\\
        \text{Site 5} &  0.023 $\pm$ 0.007  & 0.357 $\pm$ 0.008 & \textbf{0.924} $\pm$ 0.009 \\
        \textbf{Mean} & 0.026 $\pm$ 0.004 & 0.329 $\pm$ 0.030 & \textbf{0.936} $\pm$ 0.017  \\
        \bottomrule
    \end{tabular}
    \begin{tablenotes}
        \small
        \item[*] Training Site
    \end{tablenotes}
    \end{threeparttable}
\end{table}

We evaluate \sysname{} in terms of {\em success rate} and {\em efficiency}. 
The success rate is defined as the proportion of search processes that successfully converge within 
a specified proximity threshold $d$ of the target tag (by default, we set $d=100\,\mbox{m}$). 
Let $Q$ denote the total number of search processes in evaluation, 
$q=1,2,\ldots,Q$ denote the index of the processes,   
and $s_{k}^{(q)}$ denote the search state at the $k$th step in the $q$th search. 
The success rate is calculated as 
\begin{equation}
    \text{Success Rate}=\frac{\sum_{q=1}^{Q}\mathbbm{1}\left(\left\lVert s_K^{(q)} \right\rVert_2<d\right)}{Q},   
\end{equation}
where one search at most has $K$ steps. 
On the other hand, we evaluate search efficiency by the ratio of the theoretically shortest path (i.e., $\lVert s_0 \rVert_1$) to 
the number of steps that \sysname{} requires to converge to the tag. For instance, the efficiency of 0.1 means that the search takes $10$ times as many steps 
as the shortest path (note that $\text{Efficiency}=1$ is theoretically impossible to achieve). 
Formally, the search efficiency is computed as
\begin{equation}
    \text{Efficiency}=\frac{1}{Q'}\sum_{q=1}^{Q'} \frac{\left\lVert s_0^{(q)} \right\rVert_1}{k^*},   
\end{equation}
where the index of the earliest converged step is 
\begin{equation}
  k^*=\min\{k\in [0,K+1]\mid \forall i\in[k,K]\colon\lVert s_i \rVert_2<d \}. 
\end{equation}
Note that $k^*=K+1$ means that the search does not succeed, and thus we do not count them in the efficiency evaluation (the number of successful searches is denoted as $Q'$).   

We compare \sysname{} with the following baseline approaches: 
\begin{itemize}[leftmargin=*]
    \item {\em Ranging method}~\cite{zhu2024emergency} uses the trilateration based on RSSI ranging. 
    In each step, it uses the search data to localize the tag (i.e., $\{u_{0:k},v_{0:k}\}\to c$) and navigates the agent toward the estimated location.
    \item {\em Nelder-Mead simplex algorithm}~\cite{gao2012implementing} is a heuristic method of non-derivative trajectory-based optimization. 
    Each action is determined based on the ``gradient'' estimated from the simplices formed in the search path. 
    \item {\em Robbins-Monro (RM) algorithm}~\cite{toulis2021proximal} is a heuristic method of stochastic approximation, which uses momentum to address the noisy RSSI. We implement it with the simplex algorithm. 
    \item {\em CMIYC (DRL)}~\cite{soorki2024catch} is the SOTA method of LoRa tag search using reinforcement learning, where 
    the RL model uses LSTM to extract temporal features from the sequence of search data. 
    In the experiment, we implement the LSTM with two hidden layers and three fully connected layers, each with 128 dimensions (to ensure sufficient model capacity), the history length of 40, 
    and other parameters are identical to ours.  
    To compare under the setting of zero-shot generalization, we implement its ``single-environment'' version. 
\end{itemize} 

Unless specified otherwise, we set the grid size to $100\times 100\,\mbox{m}^2$ and the maximum steps as $K=500$. 
The action classifier is optimized by Adam optimizer with the learning rate of $1e-3$, the batch size of $500$, and the number of epochs set to $4,000$. 
The RSSI expectation at the tag location is set to $-80$. 

\begin{figure*}[tp]
  \centering
  \begin{minipage}{0.32\linewidth}
    \centering
    \includegraphics[width=\linewidth]{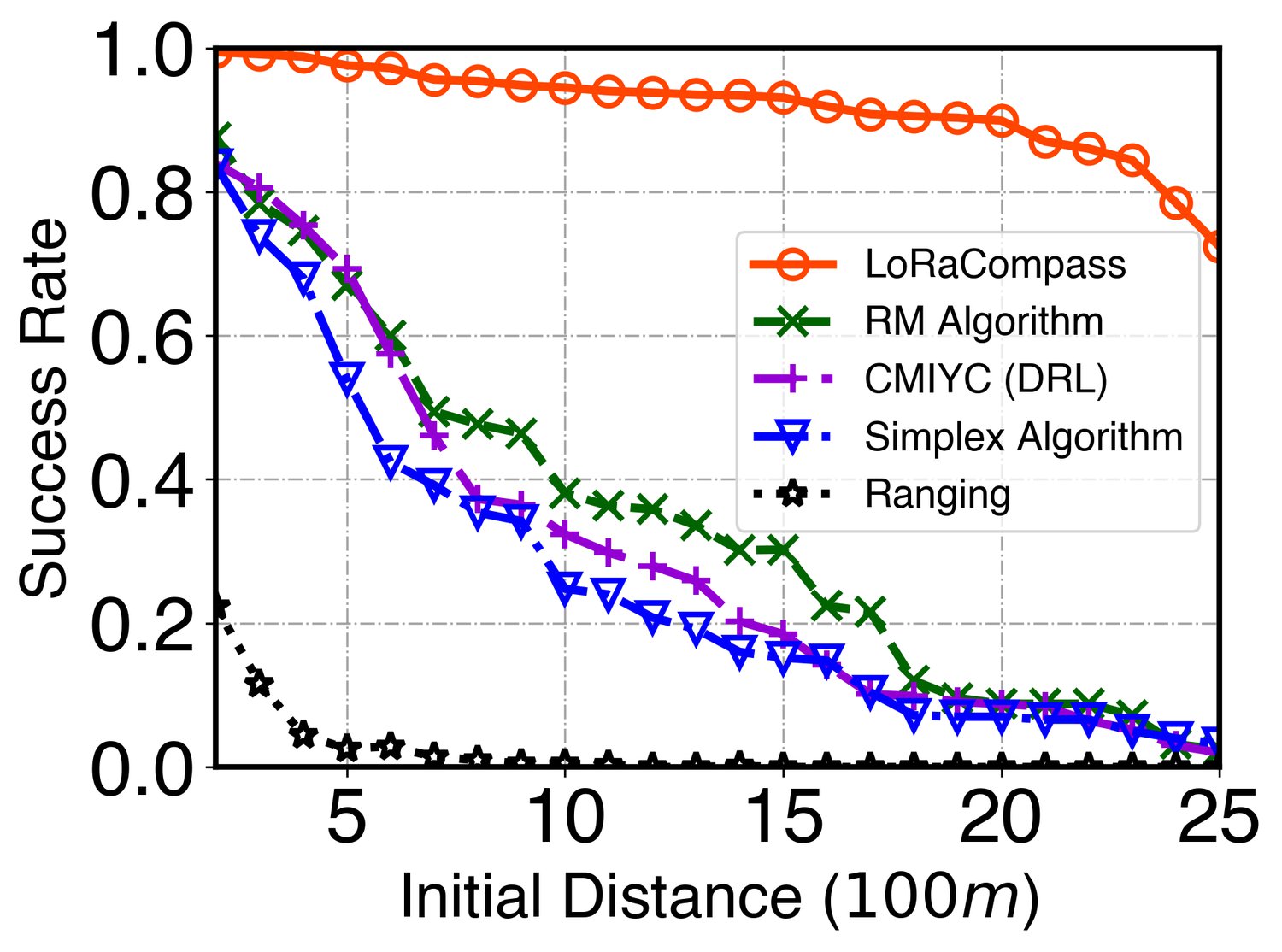}
    \caption{Success rate versus initial distance.}
    \label{fig:exp_sr_dist}
  \end{minipage}\hfill
\begin{minipage}{0.32\linewidth}
    \centering
    \includegraphics[width=\linewidth]{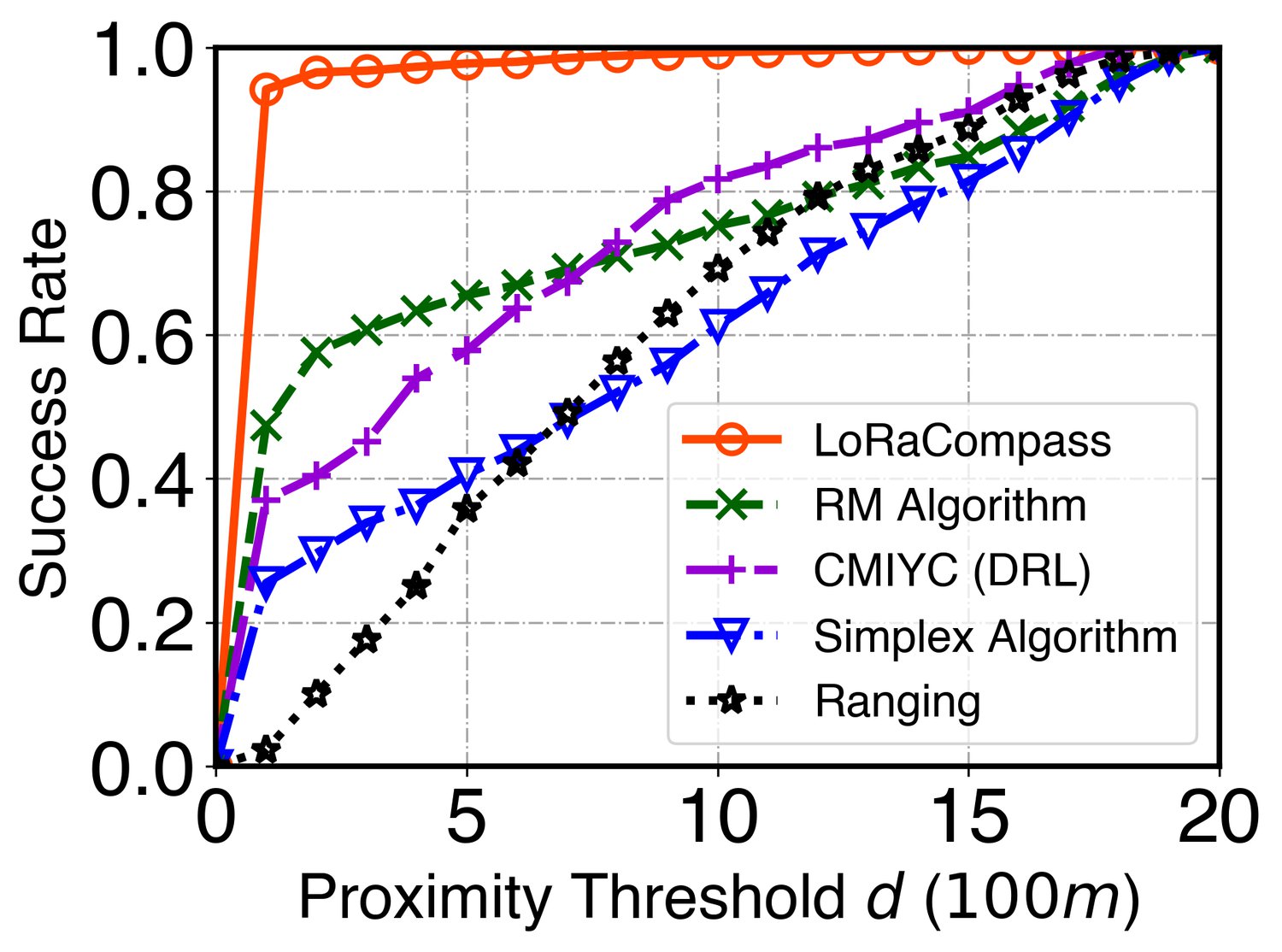}
    \caption{Success rate versus proximity threshold (i.e., localization error).}
    \label{fig:exp_se_dist}
  \end{minipage}\hfill
  \begin{minipage}{0.3\linewidth}
    \centering
    \includegraphics[width=\linewidth]{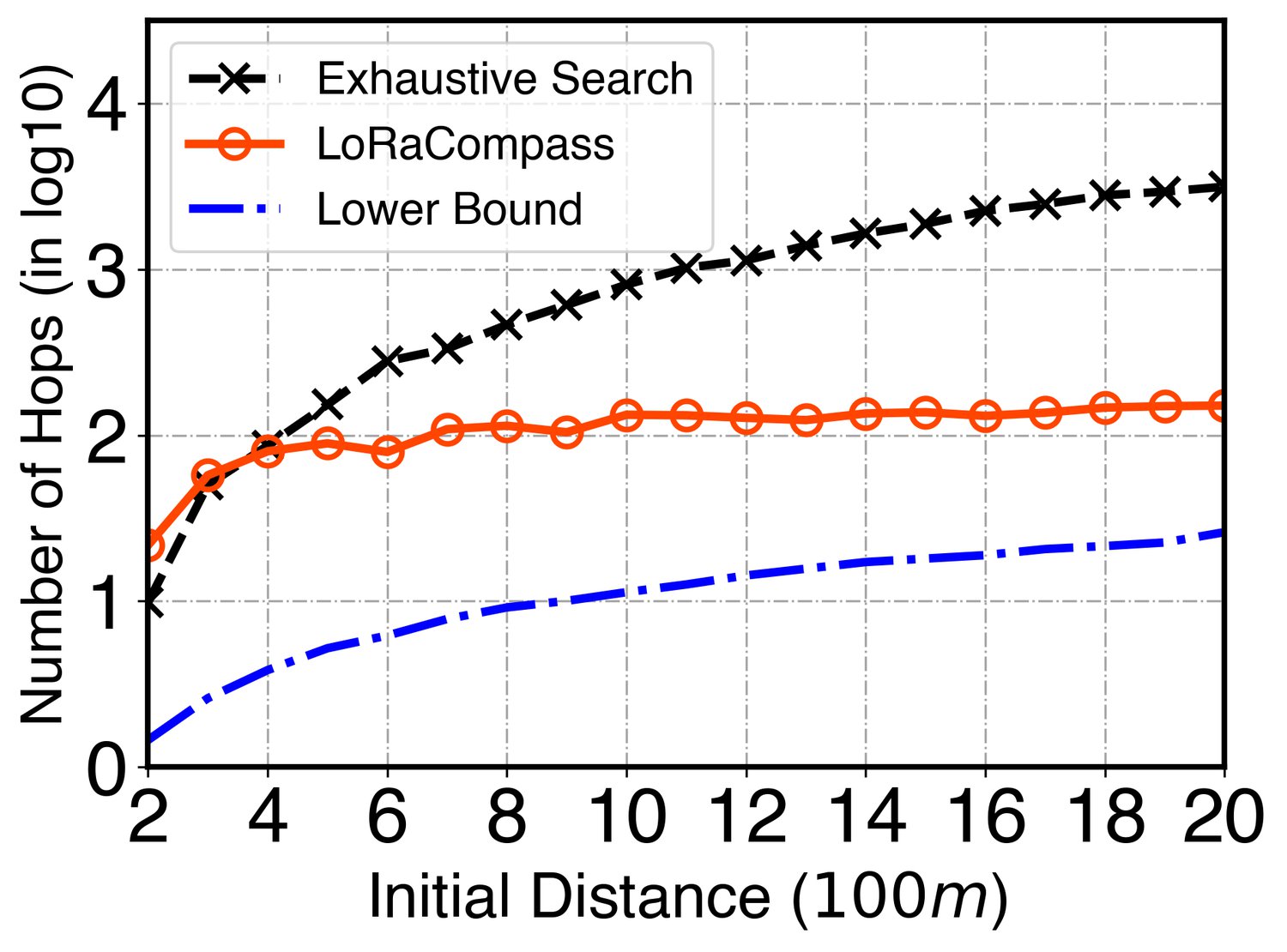}
    \caption{Number of hops versus initial distance.}
    \label{fig:exp_step_dist}
  \end{minipage}
\end{figure*}
\begin{figure*}[tp]
  \centering
  \begin{minipage}{0.24\linewidth}
    \centering
    \includegraphics[width=\linewidth]{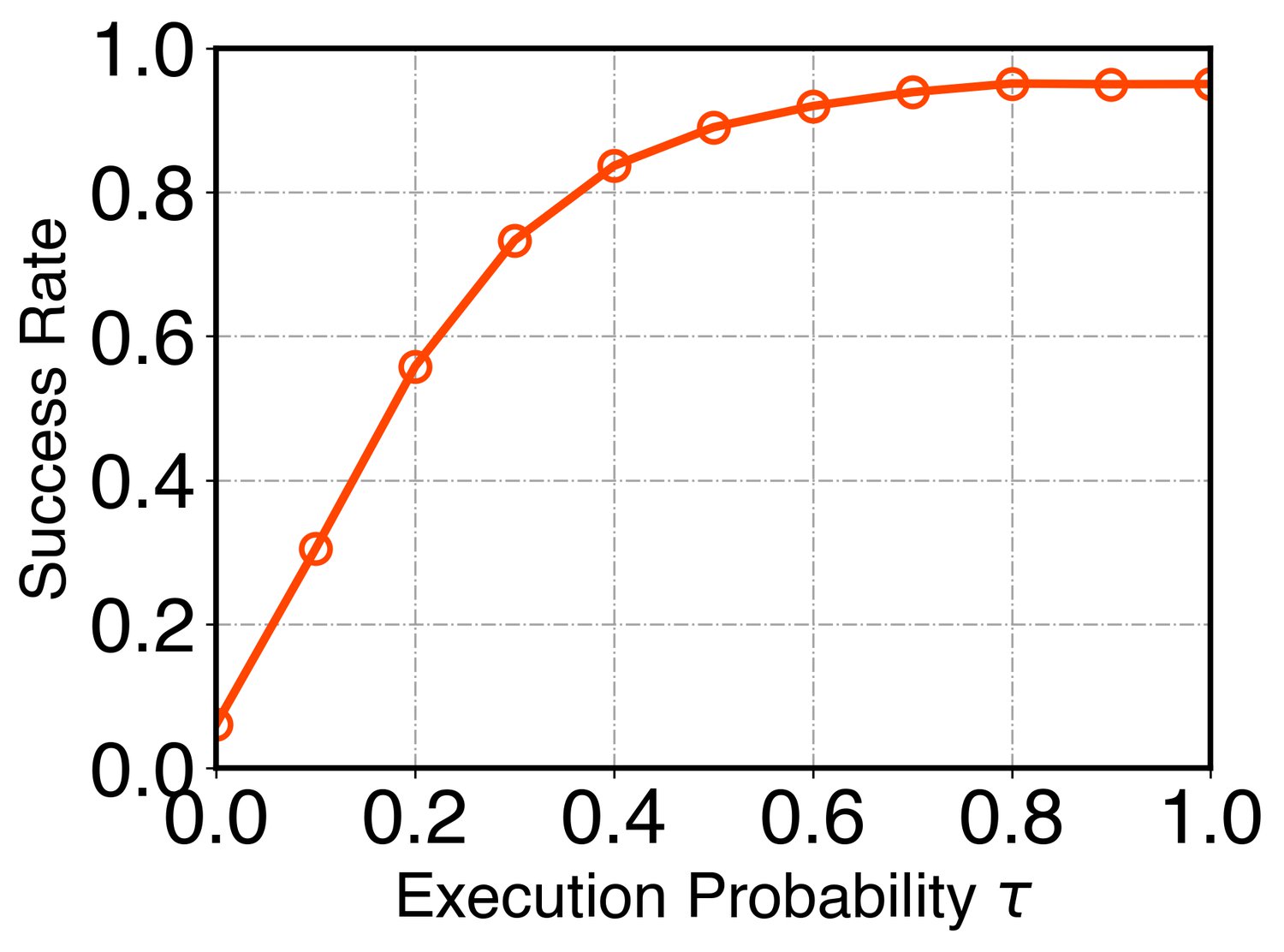}
    \caption{Success rate under actuation uncertainty.}
    \label{fig:exp_sr_rand}
  \end{minipage}\hfill
  \begin{minipage}{0.24\linewidth}
    \centering
    \includegraphics[width=\linewidth]{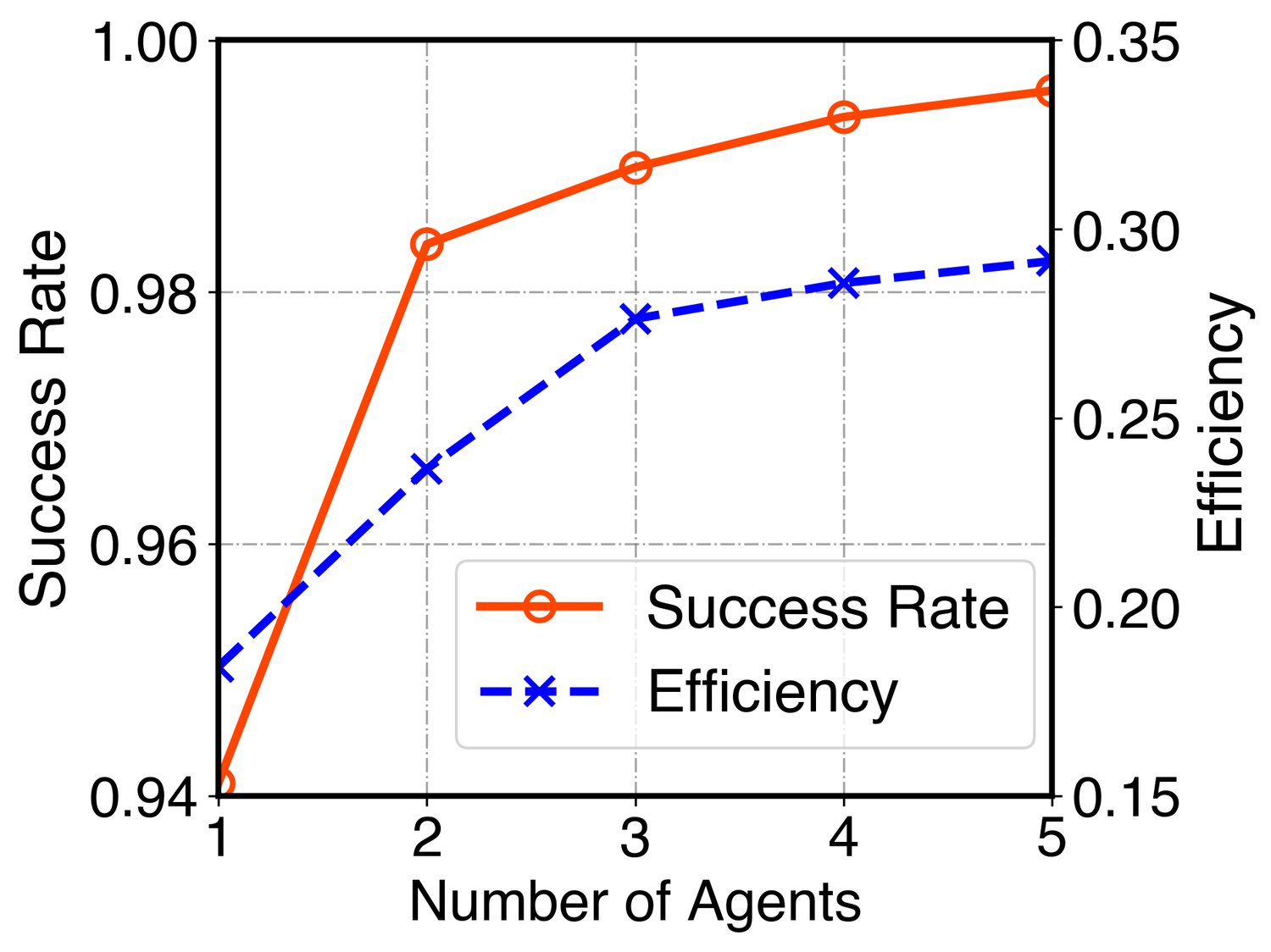}
    \caption{Performance in multi-agent search. }
    \label{fig:exp_sr_sub_n}
  \end{minipage}\hfill
  \begin{minipage}{0.24\linewidth}
    \centering
    \includegraphics[width=\linewidth]{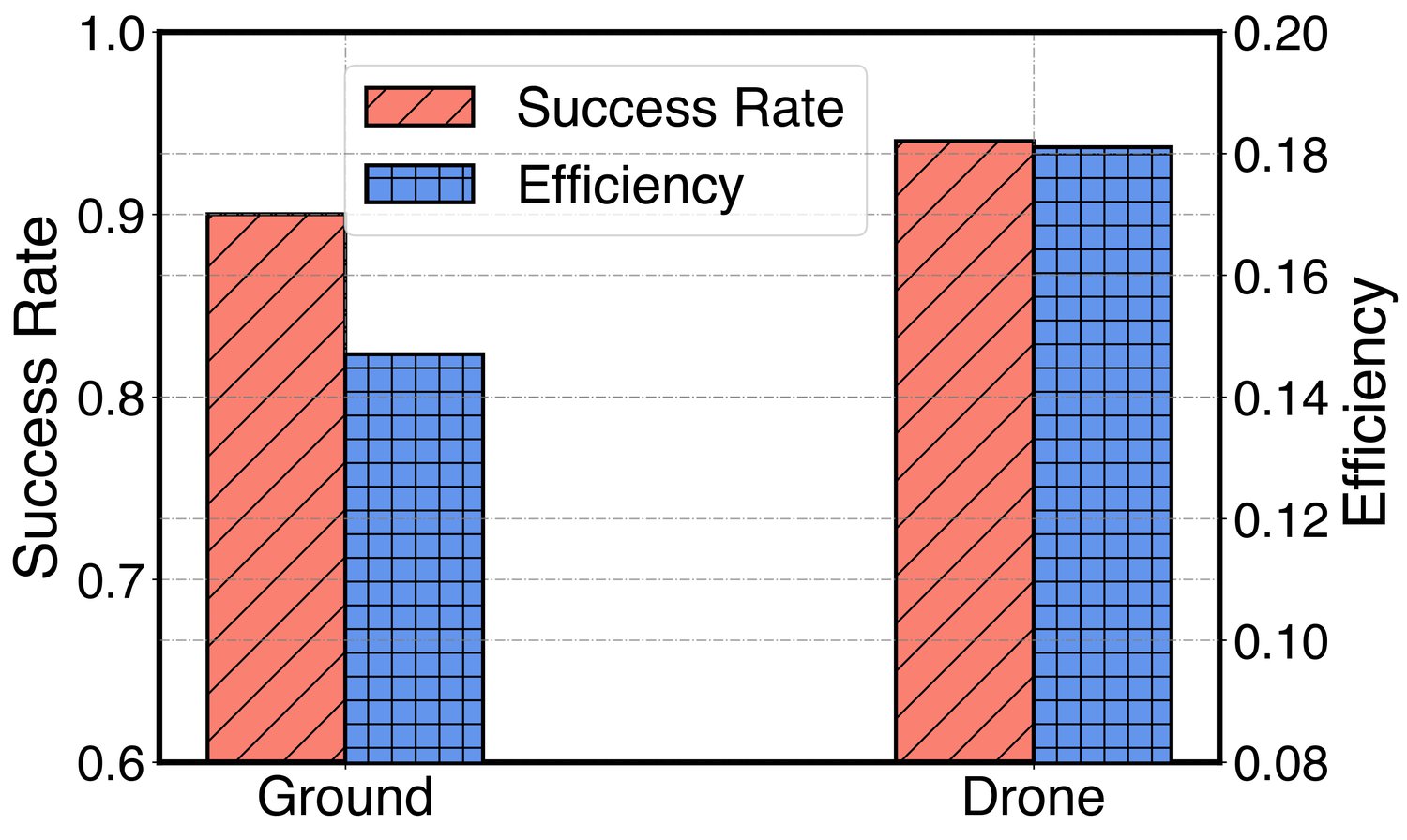}
    \caption{Performance in ground-based and drone-assisted search.}
    \label{fig:exp_deploy}
  \end{minipage}\hfill
  \begin{minipage}{0.23\linewidth}
    \centering
    \includegraphics[width=\linewidth]{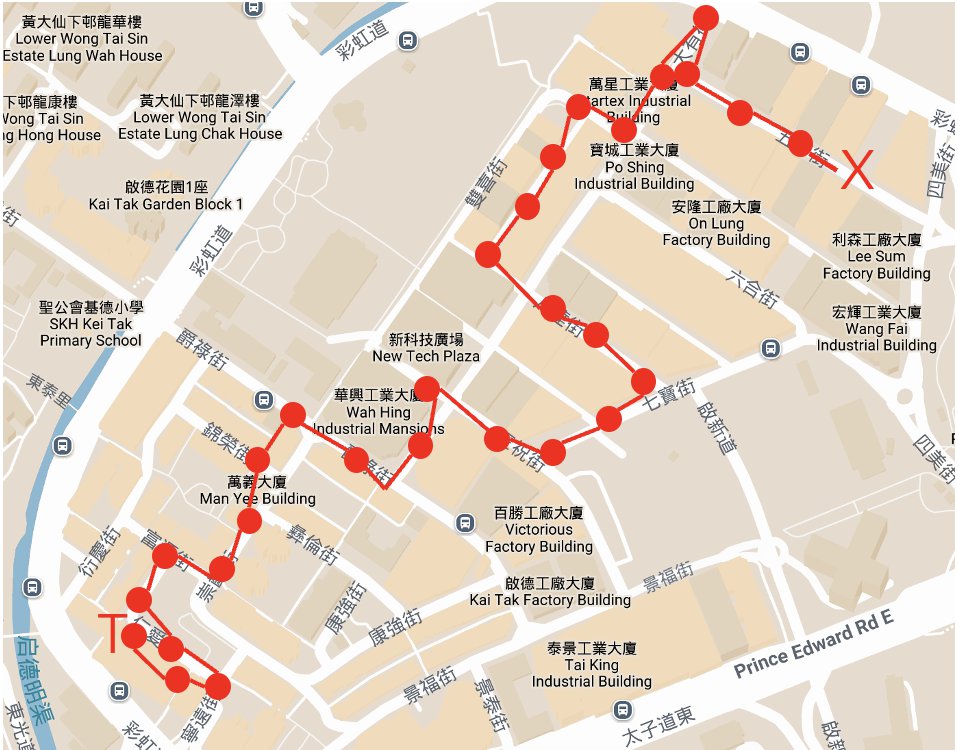}
    \caption{Search path in ground search. }
    \label{fig:exp_map}
  \end{minipage}
\end{figure*}

\subsection{System Performance}\label{subsec:exp_sim}

Table~\ref{table:sr_1} presents the overall success rates of all compared methods, where models are trained on `Site 1' and evaluated on the remaining sites. Results are reported for initial distances within 
$2\,\mbox{km}$, aligning with the typical coverage range of LoRa signals. 
Among the baselines, the ranging-based method fails in most trials due to the unreliability of 
RSSI-derived distance estimates in noisy environments, 
rendering it unsuitable for LoRa tag search. The simplex method improves upon ranging by formulating the task 
as a direct search problem, while the RM method further enhances robustness by accounting for RSSI noise. 
Nevertheless, both methods struggle to adapt to the complex, environment-dependent RSSI landscape, 
resulting in suboptimal success rates. 
The DRL-based method, CMIYC, achieves strong performance on the training site, 
outperforming all rule-based baselines. 
However, its reliance on temporal RSSI features makes it vulnerable to fluctuation and domain shifts, 
causing a notable drop in performance when tested on unseen sites.
In contrast, \sysname{} leverages spatial RSSI features to learn a generalizable representation across environments. 
It consistently achieves a high success rate ($\textgreater90\%$) across all five sites, 
demonstrating robust localization performance against RSSI fluctuation and domain shift.

\TableWord~\ref{table:sr_2} reports the success rates when all models are trained on ``Site 2''. 
With a different training site, \sysname{} maintains a high success rate, consistent with the results in \TableWord~\ref{table:sr_1}. 
(Note that the simplex and RM methods are omitted, as their performance is independent of the training site.)
This consistency arises because \sysname{} leverages the RSSI spatial pattern, which is a generalizable feature present across all sites. 
These results validate the generality and context-efficiency of \sysname{}, confirming its robust performance regardless of the training site.

\FigureWord~\ref{fig:exp_sr_dist} presents the success rates of the compared schemes over the initial distance between the agent and the tag. 
\sysname{} maintains a high success rate exceeding $90\%$ for distances up to $2\,\mbox{km}$, beyond which the rate begins to decline. 
This degradation is primarily attributed to frequent signal loss at longer ranges, under which the search behavior gradually degenerates into a random walk.

\FigureWord~\ref{fig:exp_se_dist} shows the success rate over the proximity threshold $d$, which can also be interpreted as the localization error. 
In the result, baseline schemes suffer from large localization errors due to the premature convergence, path divergence, and path looping in the search process. 
In comparison, \sysname{} improves the generalization ability, 
achieving a success rate of $94\%$ at a threshold of $100\,\text{m}$, and the rate increases to $98\%$ at $500\,\text{m}$.
These results demonstrate the satisfactory localization accuracy of \sysname{}.

\FigureWord~\ref{fig:exp_step_dist} presents the number of hops required by \sysname{} to locate a tag under varying initial distances (on a $\log_{10}$ scale). 
We use exhaustive search and the shortest path (achievable only with ground-truth information) as references.
As shown in the figure, the exhaustive search grows quadratically with the initial distance, making it highly inefficient.
In contrast, \sysname{} exhibits a near-linear trend, closely resembling the theoretical shortest path, which validates the efficiency of \sysname{}.

\FigureWord~\ref{fig:exp_sr_rand} demonstrates the robustness of \sysname{} under actuation uncertainty. 
Recall that $\tau$ denotes the probability that a chosen action is actually executed (see \SectionWord~\ref{subsec:prob_overview}). 
As shown in the figure, \sysname{} maintains a success rate exceeding $90\%$ when $\tau\textless 0.5$. 
This impressive robustness stems from the map representation, robust exploitation, and effective exploration, which collectively decouple decision-making from the action sequence.

\FigureWord~\ref{fig:exp_sr_sub_n} illustrates the search results using multiple agents. 
All agents start from the same location and search independently, while sharing a common signal map.
In this multi-agent setting, a search is considered successful once any agent locates the tag.
As shown in the figure, both success rate and efficiency improve with the number of agents; with only three agents, the system achieves a success rate exceeding $98\%$ and an efficiency score of 0.3.
This result highlights the potential for further performance gains through multi-agent collaboration.

\FigureWord~\ref{fig:exp_deploy} presents the performance of \sysname{} in real-world experiments. 
The ground-based search exhibits a slight decrease in efficiency due to increased actuation uncertainty, whereas the drone's flight path is less restricted. 
Despite this, both scenarios achieve satisfactory performance, consistent with the results from the real-data experiments. 
These findings validate the practicability of \sysname{} in real-world deployment.

Finally, we visualize the path of one ground search in \FigureWord~\ref{fig:exp_map}. 
The agent starts from $X$ and searches for the tag located at $T$. 
Moving at a normal walking speed, the agent covers a travel distance of approximately $100\,\text{m}$ per hop (marked by red dots). 
When constrained by the map, the agent selects the best feasible action according to \sysname{}'s instructions. 
As shown in the figure, despite significant signal noise and a loss rate of around $10\%$, 
\sysname{} successfully guides the agent to the vicinity of the tag, functioning like a ``compass.'' 
This demonstrates the robustness and efficiency of \sysname{} in practical deployment.

\subsection{Parameter Study}\label{subsec:exp_param}

\begin{figure*}[tp]
  \centering
\begin{minipage}{0.24\linewidth}
    \centering
    \includegraphics[width=\linewidth]{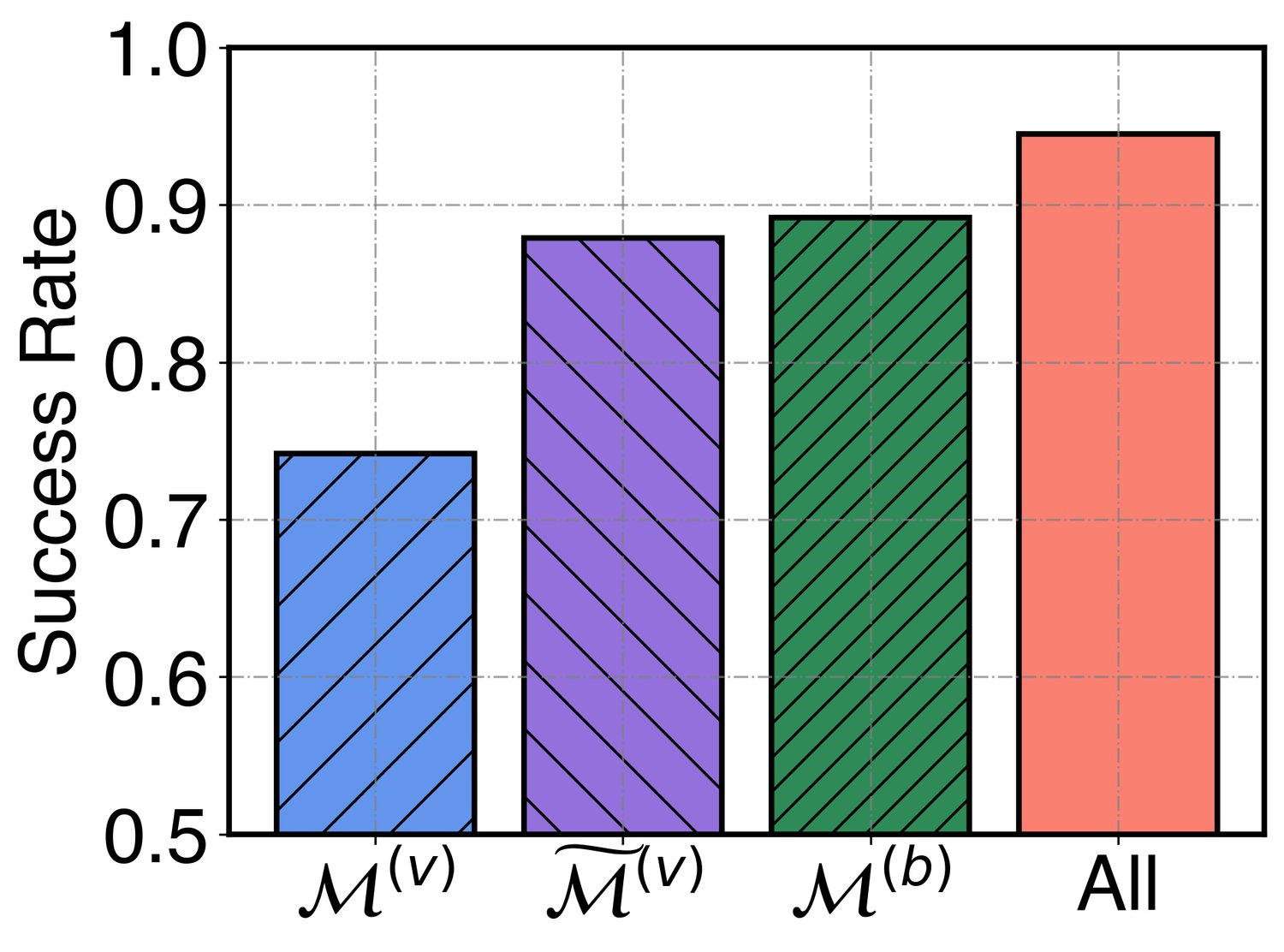}
    \caption{Ablation study on signal map.}
    \label{fig:exp_ablation_study_map}
  \end{minipage}\hfill
   \begin{minipage}{0.24\linewidth}
    \centering
    \includegraphics[width=\linewidth]{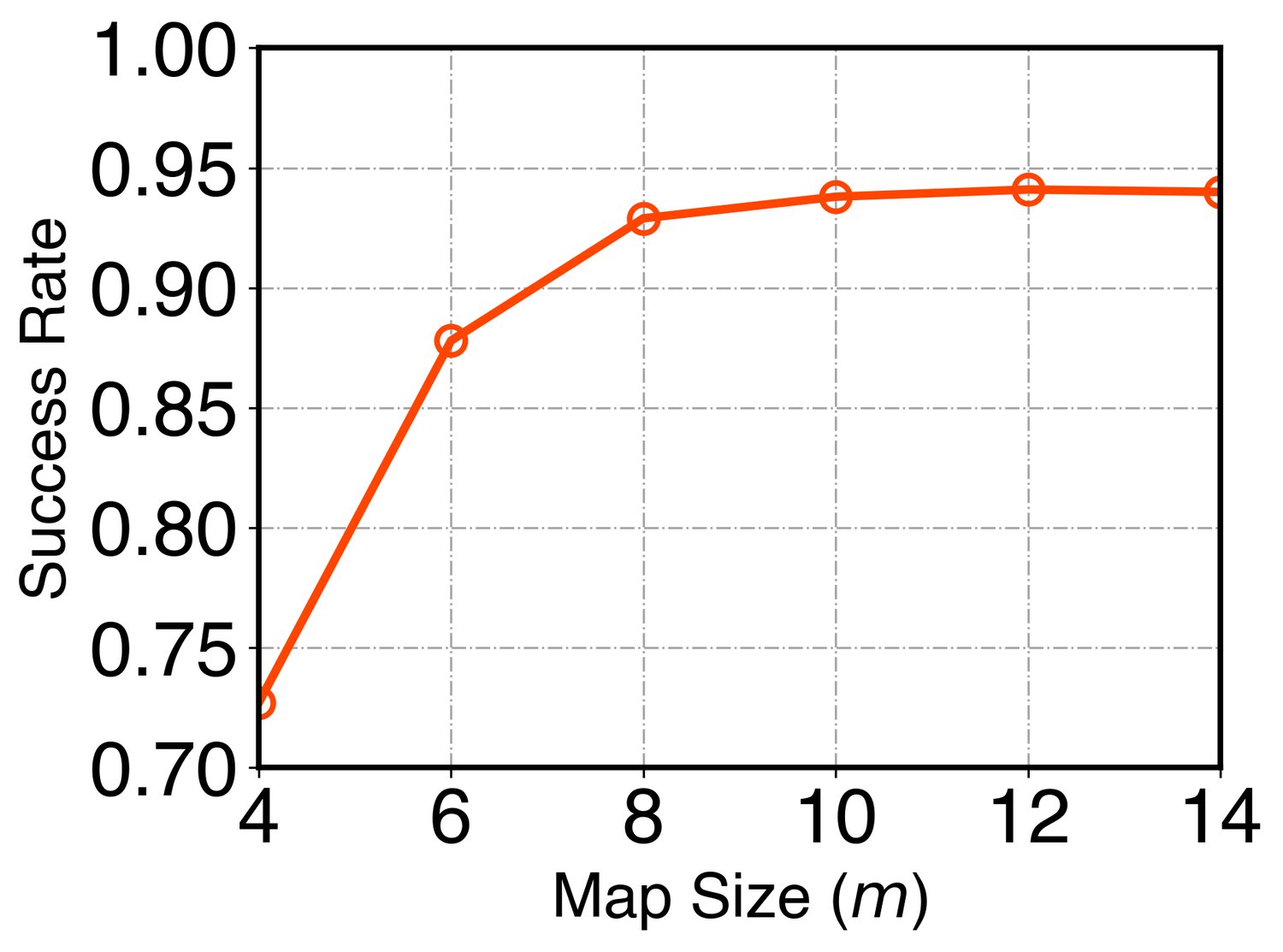}
    \caption{Parameter study on map size.}
    \label{fig:exp_param_study_perc}
  \end{minipage}
  \begin{minipage}{0.23\linewidth}
    \centering
    \includegraphics[width=\linewidth]{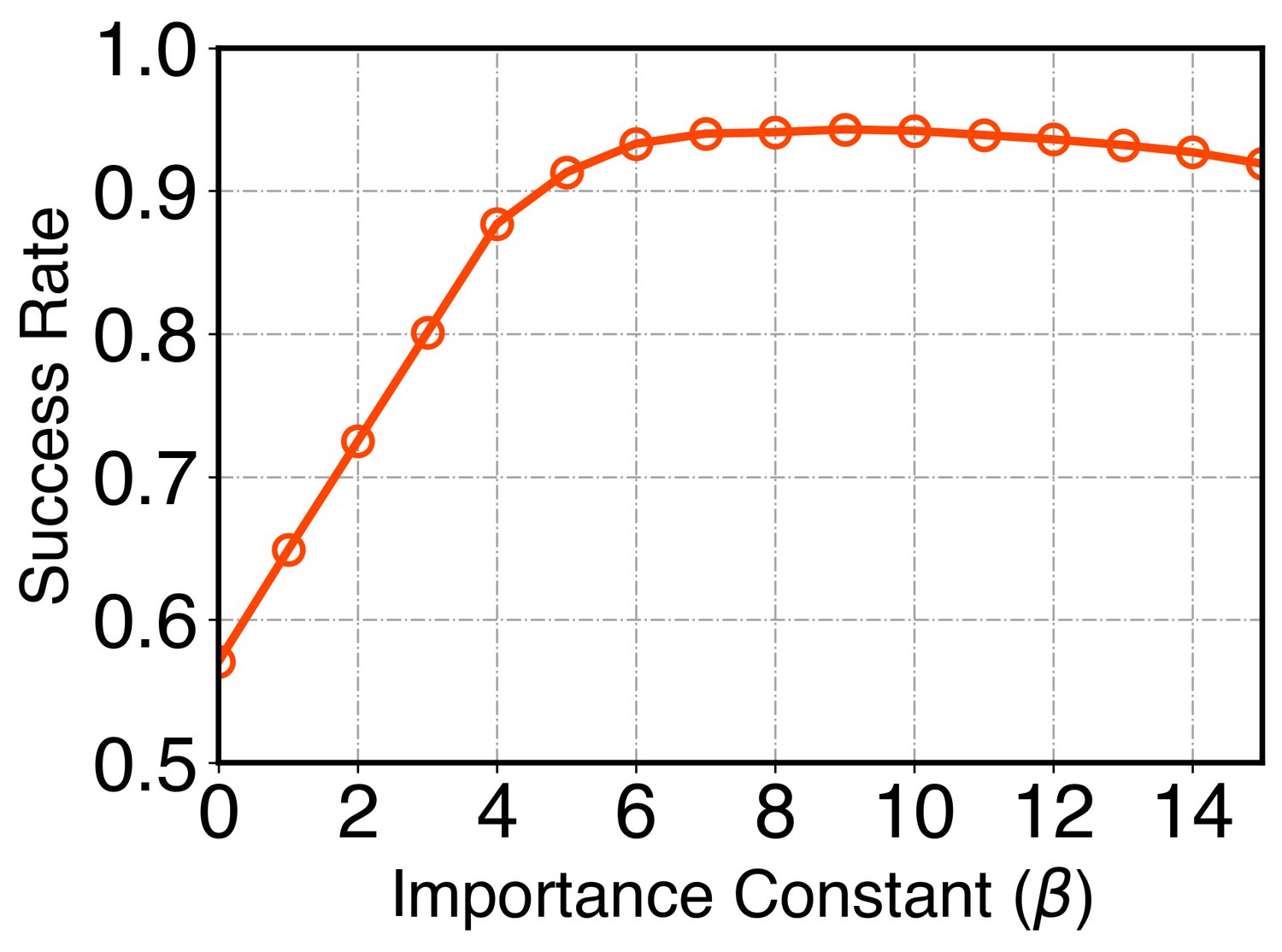}
    \caption{Parameter study on exploration importance.}
    \label{fig:exp_param_study_ucbw}
  \end{minipage}
  \hfill
  \begin{minipage}{0.24\linewidth}
    \centering
    \includegraphics[width=\linewidth]{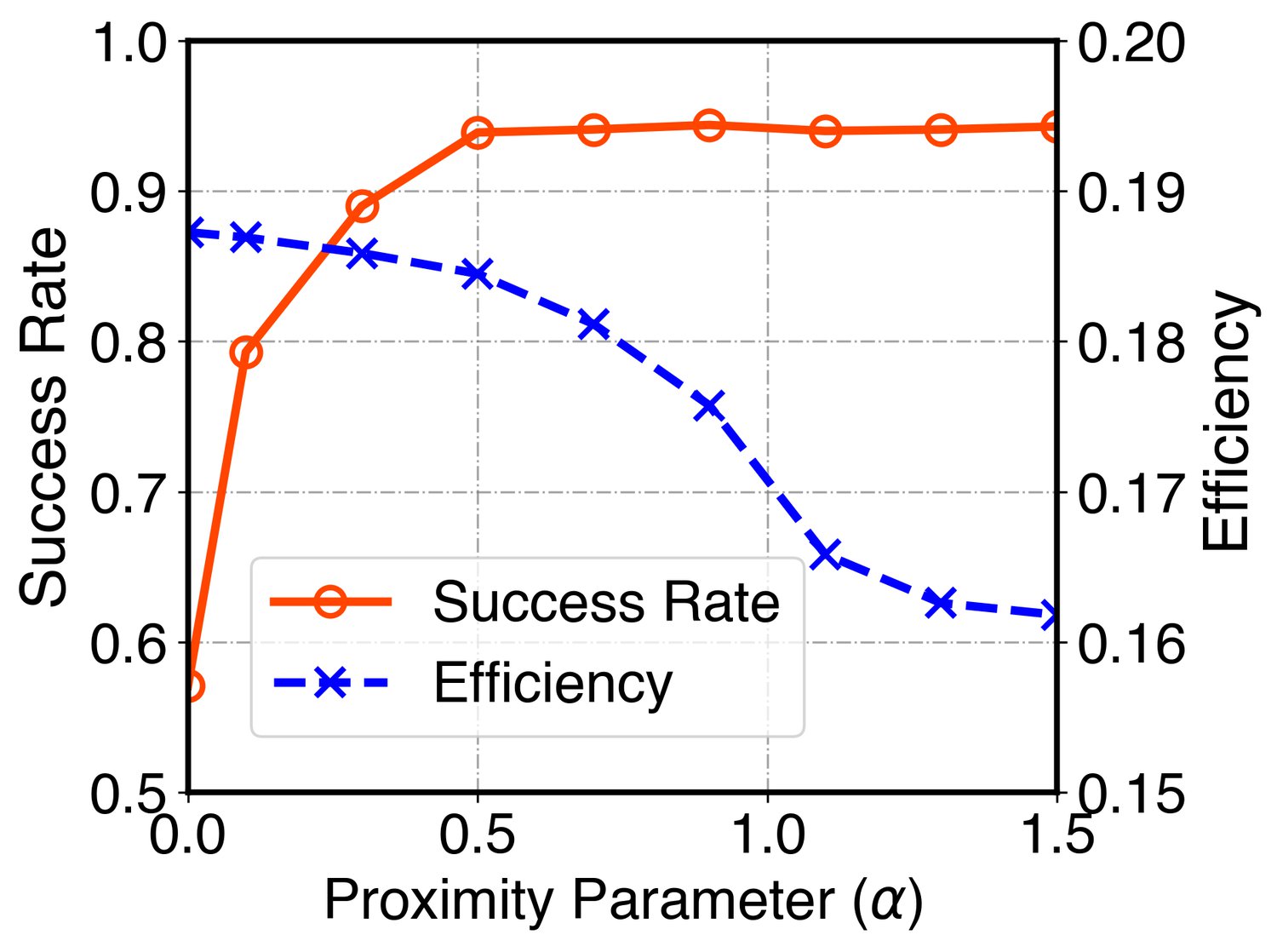}
    \caption{Parameter study on proximity parameter.}
    \label{fig:exp_param_study_alpha}
  \end{minipage}
\end{figure*}
\begin{figure*}[tp]
  \centering
  \begin{minipage}{0.33\linewidth}
    \centering
    \includegraphics[width=0.9\linewidth]{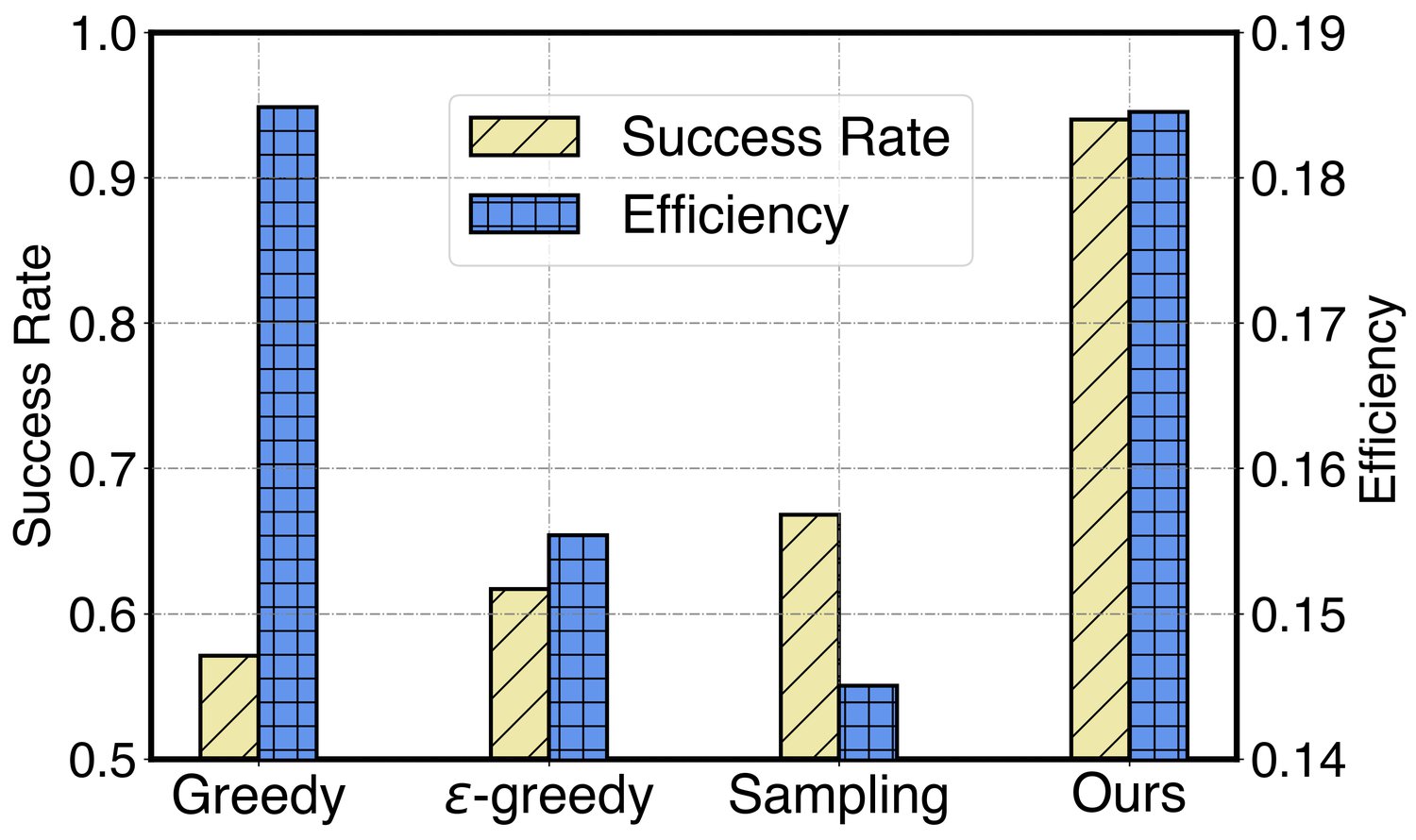}
    \caption{Comparison of exploration function. }
    \label{fig:exp_comp_explore}
  \end{minipage}\hfill
   \begin{minipage}{0.33\linewidth}
    \centering
    \includegraphics[width=0.76\linewidth]{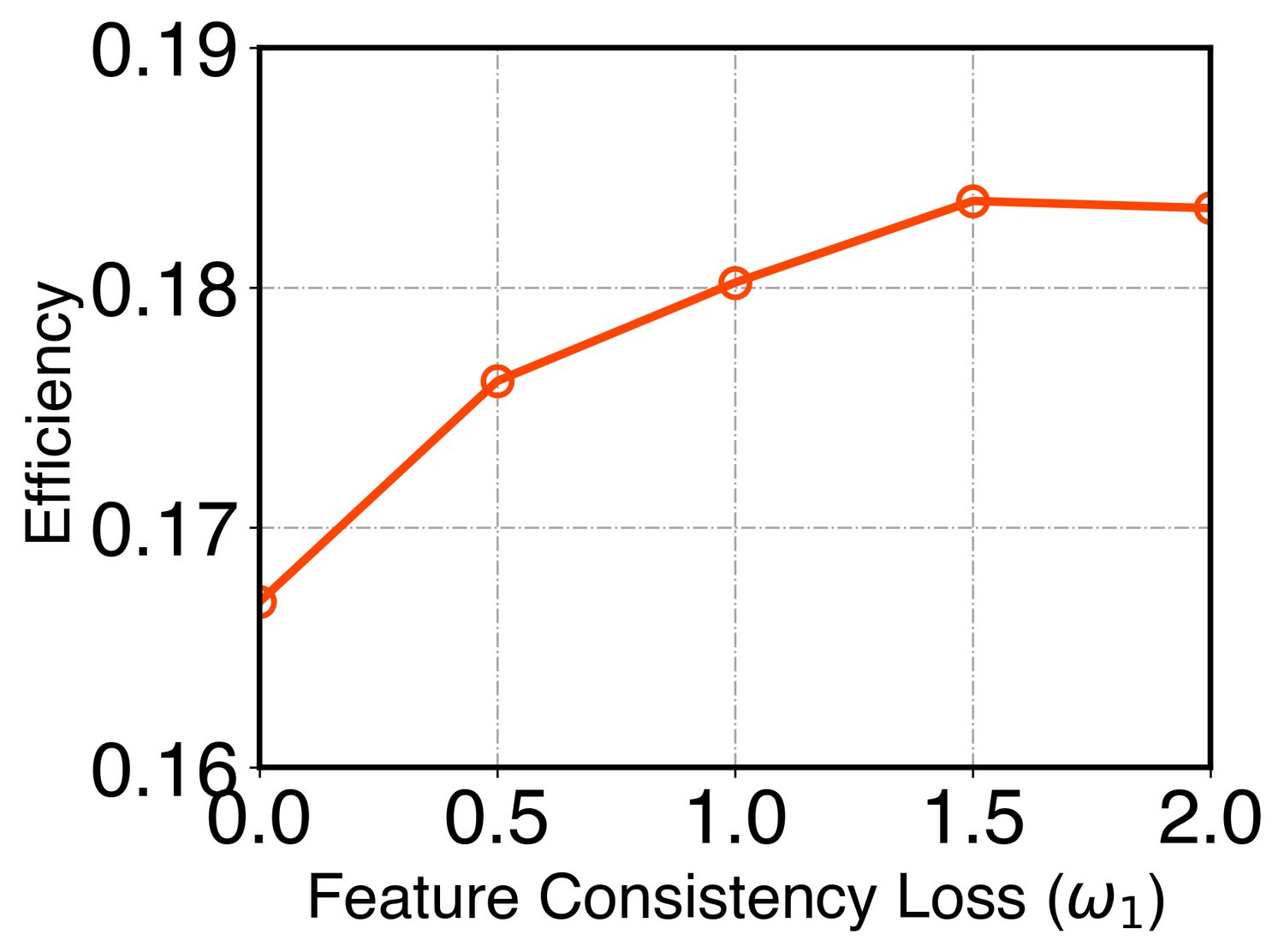}
    \caption{Study on feature-consistency loss function.}
    \label{fig:exp_param_study_pd}
  \end{minipage}\hfill
  \begin{minipage}{0.33\linewidth}
    \centering
    \includegraphics[width=0.85\linewidth]{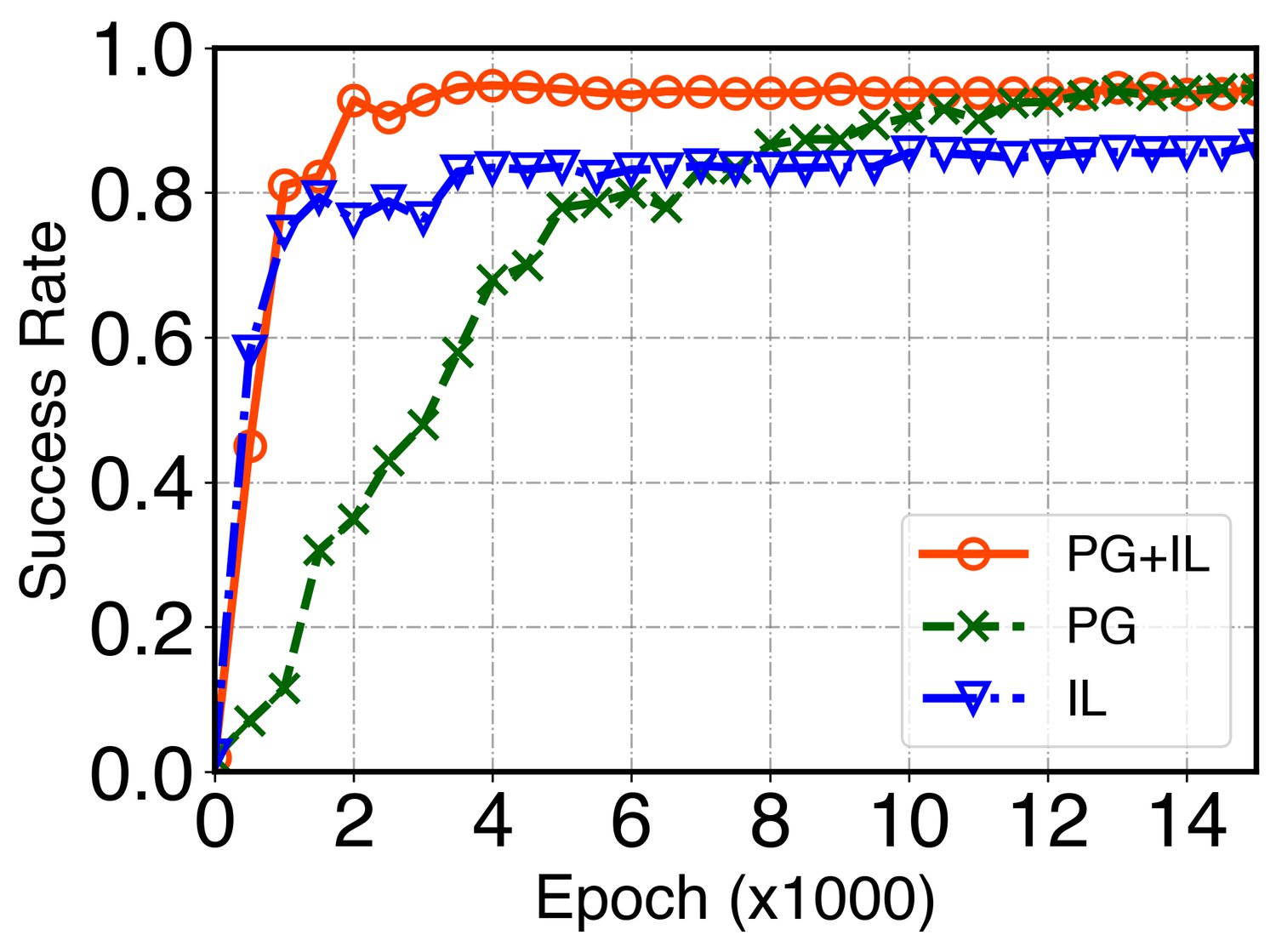}
    \caption{Study on imitation-learning loss function ($\omega_2=1$).}
    \label{fig:exp_training_eff}
  \end{minipage}
\end{figure*}

\FigureWord~\ref{fig:exp_ablation_study_map} presents the success rate when components of the signal map are in turn omitted. 
Without $\mathcal{M}^{(v)}$, the agent fails to maintain a high success rate due to the loss of critical information. 
Removing $\widetilde{\mathcal{M}}^{(v)}$ leads to a decline in performance compared to the full signal map, as the relationship between the tag direction and RSSI increase is no longer emphasized. 
Additionally, $\mathcal{M}^{(b)}$ provides visiting history; without it, the search cannot distinguish sampled signals from data imputation. 
Collectively, these results demonstrate that all three components are essential for achieving a high success rate.

\FigureWord~\ref{fig:exp_param_study_perc} shows a parameter study on the map size, controlled by $m$.  
In the figure, the success rate increases with $m$ and saturates beyond $m=10$, meaning that this size is sufficient for stable performance. Thus, we set
$m=10$ in our experiments. 

\FigureWord~\ref{fig:exp_param_study_ucbw} presents a parameter study on the exploration importance constant $\beta$. 
The success rate increases with $\beta$, reaches a plateau at $\beta=8$, and slightly decreases after $\beta=13$. 
This behavior reflects the role of $\beta$ in balancing exploration and exploitation: a small value fails to resolve looping, while an excessively large value may undermine the exploitation capability. 
Based on these observations, we set $\beta=8$ in our experiments.

\FigureWord~\ref{fig:exp_param_study_alpha} presents a parameter study on the proximity parameter $\alpha$, which controls the soft boundary for confidence tuning (see \EquationWord~(\ref{eq:sigma_function})). 
As shown in the figure, a very small $\alpha$ leads to premature convergence and thus large localization errors, while an excessively large $\alpha$ increases the number of hops required for convergence, reducing efficiency. 
To balance these two effects, we set $\alpha=0.5$ in our experiments.

In \FigureWord~\ref{fig:exp_comp_explore}, we compare our exploration function with other common exploration strategies: the greedy method selects the action with the highest hop-closer probability; $\epsilon$-greedy selects a random action with probability $\epsilon$; and the sampling method directly samples from the hop-closer probability distribution. 
As shown in the figure, the greedy method suffers from path looping, resulting in a low success rate. While both $\epsilon$-greedy and the sampling method alleviate looping to some extent, they introduce randomness that trades off efficiency. 
In contrast, our exploration function optimizes map completeness to address path looping without introducing randomness, thereby achieving both high success rate and search efficiency.

\FigureWord~\ref{fig:exp_param_study_pd} presents the parameter study on the weight $\omega_1$ of the feature-consistency loss. 
As shown in the figure, search efficiency improves with increasing $\omega_1$, as it helps mitigate signal fluctuation. 
However, an excessively large weight may restrict the flexibility of the learned features. 
To balance this trade-off, we set $\omega_1 = 1.5$ in our experiments.

\FigureWord~\ref{fig:exp_training_eff} presents the training curves with ($\omega_2=1$) and without ($\omega_2=0$) the imitation learning (IL) loss function. 
As shown in the figure, using only the policy gradient (PG) loss achieves good performance, but convergence is slow. 
In contrast, the IL loss converges quickly, yet its final performance is unsatisfactory. 
By combining both losses, our method achieves strong search performance with efficient training convergence.

\section{Related Work}\label{sec:related_work}

Existing technologies locate a LoRa tag using either a fixed gateway infrastructure or a mobile sensor~\cite{aldhaheri2024lora, aernouts2018sigfox, li2021urban, guo2022illoc}. 
The former conducts trilateration or fingerprinting based on the LoRa gateways installed as civil infrastructure~\cite{aernouts2018sigfox,li2021urban, lutakamale2024rssi, svertoka2022lorawan}. 
The latter works by sequentially moving a LoRa sensor toward a tag~\cite{soorki2024catch, shi2024enable, marquez2023understanding}. 
Recently, the mobile scheme has gained increasing traction, as it operates independently of LoRa infrastructure, so   
its applicability and localization accuracy are not subject to the number of LoRa gateways pre-installed. 

In the scope of source seeking, 
LoRa localization using a mobile sensor has two problem formulations:  
some formulate the problem as a localization task, while others regard it as a sequential decision-making process~\cite{shen2022cloud, wang2021multi}. 
The localization method leverages ranging technologies (AoA, ToF, TDoA, etc.) to directly locate a LoRa tag, toward which the mobile sensor can move~\cite{sorbelli2020measurement, bai2021autonomous, marquez2023understanding, toulis2021proximal, gao2012implementing, yuan2022uav, zhang2023distributed}. 
However, the ranging process is often vulnerable to real-world multipath fading and non-line-of-sight (NLOS) scenarios, making the localization methods inaccurate in the wide LoRa coverage. 
On the other hand, the decision-making method determines search directions for the mobile sensor to sequentially move toward a tag, 
using the RSSI sampled at the sensor (RSSI is widely adopted, as it is easy to obtain and suitable for NLOS scenarios)~\cite{gao2012implementing, soorki2024catch, yoshitome2022lora, teng2025secure}. 
In contrast to direct localization, the decision-making method can potentially achieve higher accuracy, because it targets the location with the maximal RSSI, where the tag is normally located. 

The decision-making for LoRa tag search can be approached by heuristic algorithms or reinforcement learning. 
The heuristic method treats the search problem as an optimization instance. 
For example, the problem can be regarded as a derivative-free trajectory-based optimization and  
tackled by the Nelder-Mead simplex algorithm~\cite{gao2012implementing, selvam2022nelder}; 
we may further account for the RSSI noise by modeling the problem as a stochastic approximation problem and apply the Robbins-Monro algorithm~\cite{toulis2021proximal}.
However, these rule-based heuristic algorithms are subject to the shape of the optimization landscape (e.g., convexity, continuity, or smoothness), making them vulnerable to the real-world complexity of LoRa propagation. 
To adapt to the real-world environments, reinforcement learning has emerged promising, with various standard models applied to this search problem~\cite{soorki2024catch, ebrahimi2020autonomous, ossongo2024multi, gu2020uav}. 
While commendable search efficiency has been demonstrated, existing methods often rely on RSSI temporal features 
that are vulnerable to RSSI fluctuation and domain shift. This often results in cascading decision errors that culminate in 
in localization inacuracies when operating across environments. 

We propose \sysname{}, the first zero-shot generalization (ZSG) reinforcement learning model that leverages the RSSI spatial feature to enable LoRa tag search robust to signal fluctuation and domain shift across environments.  
We propose a novel spatially-aware exploitation model to enhance search robustness and  
a UCB-inspired exploration function to account for search uncertainty. 
Altogether, \sysname{} allows the mobile sensor to sequentially move toward a LoRa tag with increasing confidence, notwithstanding domain shift and signal fluctuation.

\section{Conclusion}\label{sec:conclude}

We study the sequential decision-making process for a mobile LoRa sensor to locate a LoRa tag in unknown environments, using the received signal strength indicator (RSSI) at the sensor, which is 
fundamental to the tracking of mentally incapacitated patients (MIPs) and others who may go missing. 
This search process is challenging due to domain shift and RSSI fluctuation, 
as the mobile sensor needs to navigate an unpredictable, largely non-convex, and noisy RSSI landscape. 
This hinders previous works from achieving accurate and efficient tag localization.   

We introduce \sysname{}, a novel zero-shot generalizable (ZSG) reinforcement learning model 
to efficiently search for a LoRa tag under real-world domain shift and RSSI fluctuation. 
For exploitation, \sysname{} employs a robust deep learning model to capture the spatial representation from RSSI time series and 
maximize the probability of the sensor hopping toward the tag. 
For exploration, it uses a UCB (upper confidence bound)-inspired function to proactively reduce uncertainty in exploitation, 
enabling the sensor to hop toward the tag with increasing confidence. 
We have validated \sysname{} in both ground-based and drone-assisted scenarios within diverse unseen environments, covering over $80\,\mbox{km}^2$.
\sysname{} achieves a high success rate of over 90\% in locating LoRa tags within $100\,\mbox{m}$, outperforming previous methods by $40\%$, with a satisfactory near-linear search efficiency with respect to initial distance.  

Our future work will focus on extending \sysname{} to multi-sensor settings and more generalized forms, 
where collaborative efforts can enhance the search performance for LoRa tags, 
paving the way for more effective applications in search and rescue operations, location-based services, and AI healthcare.

\bibliographystyle{IEEEtran}
\bibliography{sample-base}

\newpage\appendix
\section*{Visualization of RSSI Heatmap}\label{sec:appendix_heatmap}

\begin{figure*}[!t]
  \centering
  \includegraphics[width=0.25\linewidth]{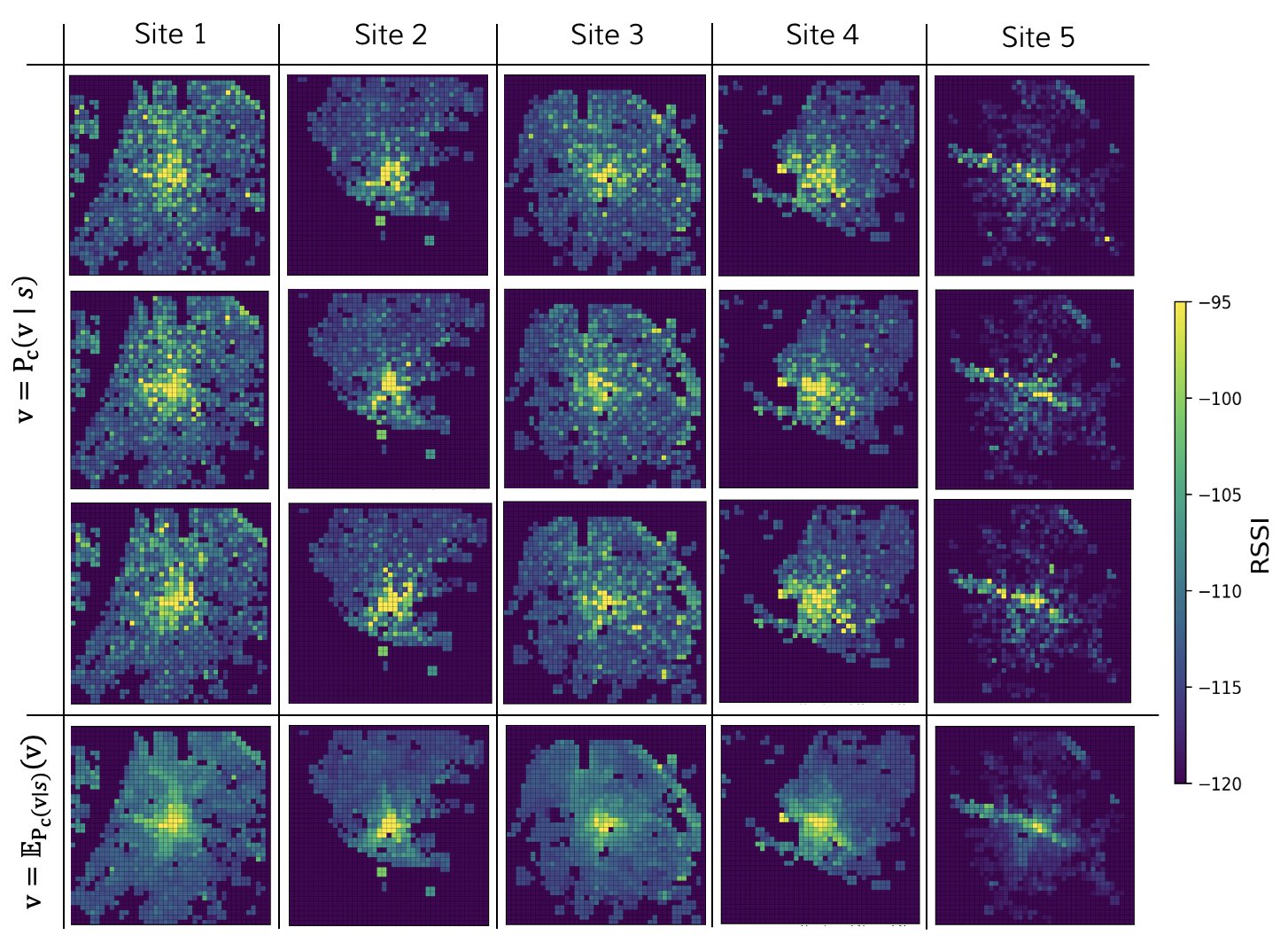}
  \caption{Illustration of RSSI heatmaps in $4.1\times 4.1km^2$. 
  An RSSI measurement $v$ is sampled from an observation function $P_c(v\mid s)$. 
  The distribution depends on the signal environments between the agent and the tag, 
  represented as the search state $s$.}
  \label{fig:heatmap_sample} 
\end{figure*}

\FigureWord~\ref{fig:heatmap_sample} illustrates the RSSI heatmaps of the five sites of $4.1\times 4.1km^2$ used in our experiment. 
The first three rows present different snapshots of the same site by sampling at different times. The last row shows 
the underlying heatmap of RSSI expectation (approximated by the mean of a large number of samples). 
In these heatmaps, significant domain shift and signal fluctuation are observed, which are due to the various environmental factors such as buildings, vegetation, terrain, and dyanmic factors like multipath fading, moving objects, and ambient noise. 
While RSSI is noisy on a local scale, they show a clear pattern from a global, macro perspective that RSSI is larger when the sensor is closer to the tag, and smaller when the sensor is farther away. 
This provides a robust representation of the tag's direction, independent of environmental variations, 
which facilitates an agent to hop to a LoRa tag.

\end{document}